\documentclass[pdflatex,sn-mathphys-num]{sn-jnl}

\usepackage{graphicx}%
\usepackage{multirow}%
\usepackage{amsmath,amssymb,amsfonts}%
\usepackage{amsthm}%
\usepackage{mathrsfs}%
\usepackage[title]{appendix}%
\usepackage{xcolor}%
\usepackage{textcomp}%
\usepackage{sistyle}%
\usepackage{manyfoot}%
\usepackage{booktabs}%
\usepackage{algorithm}%
\usepackage{algorithmicx}%
\usepackage{algpseudocode}%
\usepackage{listings}%
\usepackage{subcaption}%
\usepackage[font=small,skip=0pt]{caption}%
\newcommand{\ts}{\textsuperscript}

\theoremstyle{thmstyleone}%
\theoremstyle{thmstyletwo}%

\theoremstyle{thmstylethree}%

\begin{document}

\title[Article Title]{Compressing high-resolution data through latent representation encoding for downscaling large-scale AI weather forecast model}


\author[2]{\fnm{Qian} \sur{Liu}}\email{liuqian@sais.com.cn}
\equalcont{These authors contributed equally to this work.}
\author[3]{\fnm{Bing} \sur{Gong}}\email{gongbing1112@gmail.com}
\equalcont{These authors contributed equally to this work.}
\author[4]{\fnm{Xiaoran} \sur{Zhuang}}\email{zxrxz3212009@163.com}
\equalcont{These authors contributed equally to this work.}
\author[1]{\fnm{Xiaohui} \sur{Zhong}}\email{x7zhong@gmail.com}
\author*[5]{\fnm{Zhiming} \sur{Kang}}\email{kangzm@cma.gov.cn}
\author*[1,2]{\fnm{Hao} \sur{Li}}\email{lihao$\_$lh@fudan.edu.cn}

\affil*[1]{\orgdiv{Artificial Intelligence Innovation and Incubation Institute}, \orgname{Fudan University}, \orgaddress{\city{Shanghai}, \postcode{200433}, \country{China}}} 
\affil[2]{\orgname{Shanghai Academy of Artificial Intelligence for Science}, \orgaddress{\city{Shanghai}, \postcode{200232}, \country{China}}}
\affil[3]{\orgdiv{Department of Electrical and Computer Engineering}, \orgname{Shanghai Normal University}, \city{Shanghai}, \postcode{200234}, \country{China}}
\affil*[4]{\orgdiv{Jiangsu Meteorological Observatory}, \city{Nanjing, Jiangsu}, \postcode{210008}, \country{China}}
\affil[5]{\orgdiv{Nanjing Meteorological Observatory}, \city{Nanjing, Jiangsu}, \postcode{210009}, \country{China}}


\abstract{
The rapid advancement of artificial intelligence (AI) in weather research has been driven by the ability to learn from large, high-dimensional datasets. However, this progress also poses significant challenges, particularly regarding the substantial costs associated with processing extensive data and the limitations of computational resources. Inspired by the Neural Image Compression (NIC) task in computer vision, this study seeks to compress weather data to address these challenges and enhance the efficiency of downstream applications. Specifically, we propose a variational autoencoder (VAE) framework tailored for compressing high-resolution datasets, specifically the High Resolution China Meteorological Administration Land Data Assimilation System (HRCLDAS) with a spatial resolution of 1 km. Our framework successfully reduced the storage size of 3 years of HRCLDAS data from 8.61 TB to just 204 GB, while preserving essential information. In addition, we demonstrated the utility of the compressed data through a downscaling task, where the model trained on the compressed dataset achieved accuracy comparable to that of the model trained on the original data. These results highlight the effectiveness and potential of the compressed data for future weather research.}

\keywords{latent representation, downscaling, AI weather forecasts, varational autoencoder}



\maketitle
\section{Introduction}\label{sec1}

Weather forecasting is crucial for society and various industries, supporting informed decision-making in areas such as agriculture, transportation, disaster management. Traditionally, numerical weather prediction (NWP)) models have been employed for this purpose, but they are computationally intensive, requiring substantial computing resources. Recent advancements in deep learning offer promising alternatives to NWP models, potentially offering faster and equally accurate forecasts \citep{bi2023accurate,li2023fuxi, lam2022graphcast}. However, the effectiveness of deep learning applications in weather and climate relies heavily on the availability of large-scale datasets. The computational demands for acquiring, storing, and managing such data frequently exceed the capabilities of researchers with only modest setups, creating significant barriers for those lacking access to high-performance computing resources and data storage.

Moreover, to ensure accurate weather forecasting, numerous super-computing and research centers around the world conduct operational weather and climate simulations multiple times daily. For example, the European Centre for Medium-Range Weather Forecasts (ECMWF) manages 230 petabytes (PB) of data and processes approximately 600 million Earth observations each day. This data volume is projected to quadruple over the next decade due to increasing spatial resolution in forecasting models \citep{klower2021compressing}. While this data growth provides more opportunities for training deep learning models, it also poses challenges as the sheer data volume can overwhelm the existing super-computing infrastructure and complicate the distribution of weather products due to limited network bandwidth. Thus, effective data compression techniques are essential.

Compressing weather data is similar to the Neural Image Compression (NIC) task in the computer vision domain, where the goal is to reduce the file size of images while maintaining acceptable quality. For example, \citet{balle2016end} employed an end-to-end convolutional neural network for image compression, while \citet{chen2022lsvc} developed a framework to compress automotive stereo videos by reducing temporal redundancy. Most of the prior work has focused on using Variational Autoencoders (VAEs) for image compression \citep{balle2018variational,cheng2020learned}. For instance, \citet{cheng2020learned} demonstrated that employing discredited Gaussian Mixture Likelihoods to parameterize the latent code distributions achieves greater accuracy than traditional entropy models. 

Although NIC based on VAEs shares similarities with weather data compression in exploring latent representation patterns, significant differences remain. Natural images, with high correlation among red, green, and blue (RGB) channels, allow traditional compression techniques to exploit these relationships effectively. In contrast, while weather data exhibits correlations among its variables, these relationships are not as straightforward correlated as nature images. Thus this further complicates the application of conventional methods. 
Additionally, the volume of weather data is substantially large. For example, the ERA5 reanalysis dataset \cite{hersbach2020era5} consists of hourly data at a spatial resolution of 31 km, with 640 $\times$ 1280 points, totaling approximately 226 terabytes (TB))]. In comparison, the High Resolution China Meteorological Administration Land Data Assimilation System (HRCLDAS) \cite{CLDAS2014,CLDAS2020} offers a regional product with a spatial resolution of 1 km, incorporating  4500 $\times$ 7000 grid points hourly. Managing such large datasets for weather applications introduces challenges for NIC in deep learning, including increased data loading times and processing requirements that can hinder training and model convergence, ultimately degrading compression performance. Thus, this study proposes a new data compression method tailored specifically to weather data to address these challenges.

Furthermore, to evaluate the effectiveness of the proposed data compression method, we employed statistical downscaling as a proof-of-concept downstream application. Despite advancements in deep learning for downscaling, the conflict between the need for extensive data and limited computational resources remains a significant challenge. For instance, \citet{leinonen2020stochastic} utilized a generative adversarial model to stochastically downscale coarse radar data to high resolution without additional conditional variables. \citet{harris2022generative} divided the entire UK region into smaller patches to incorporate more conditional inputs given computational constraints. yet this approach limits the deep learning model's ability to capture global features. Similarly, \citet{zhong2024investigating} used high-resolution assimilation data as ground truth for downscaling, but their application was confined to specific small regions in China. Thus, demonstrating the successful application of compressed data for these tasks could serve as a promising example in the field, alleviating computational constraints and promoting broader application of deep learning techniques in weather downscaling.

In addressing the challenges of managing large-scale weather data, we propose a NIC framework specifically designed for this purpose, compressing high-resolution weather data into a latent representation space to ease the data burden for downstream tasks. The main contributions of this research are as follows:

\begin{itemize}
    \item \textbf{Introduction of a novel data compression framework tailor to weather data:} We present a Variational Autoencoder (VAE) method for reducing high-resolution weather data. The VAE's encoder generates a quantize latent Gaussian distribution through the variational inference process. Our comparison of training strategies reveals that pre-training followed by a fine-tuning yields the best reconstruction performance, resulting in superior latent representation of the original data.

    \item \textbf{Proof of Concept with large HRCLDAS Data}: We applied our framework to 1-km high-resolution HRCLDAS data, successfully compressing three years' worth of data from a total of \textbf{8.61 TB} to a compact \textbf{204 GB} data. Reconstruction results demonstrate that the VAE decoder can accurately reproduce the raw data with minimal loss of information and detail, effectively recovering and preserving key properties of the original data, such as extreme values.
    
    \item \textbf{Downstream Application - statistical downscaling:} We provide a downstream application - statistical downscaling the output from a deep learning based weather forecasting model, i.e. FuXi \cite{chen2023fuxi},  over China as proof of concept. Unlike most studies that resolve the NWP outputs, our research is the first to downscale deep learning-based weather forecasting model outputs, enhancing their resolution. Our results indicate that our framework significantly reduces computational costs while maintaining model performance. 
    

\end{itemize}

\section{Methodology}\label{sec2}

\subsection{Data sources}\label{sec:data_source}


To assess the performance of the proposed compression framework, we employed downscaling as a proof-of-concept application, using HRCLDAS as the high-resolution ground truth. HRCLDAS \cite{CLDAS2014,CLDAS2020} is a blended dataset that integrates station observations, satellite data, and NWP data station advanced land surface and data assimilation techniques \citep{han2020evaluation,zhong2024investigating}. In this study, the HRCLDAS data encompasses three years of data from 2019 to 2021, including hourly 2-meter temperature (T$_{2M}$), 10-meter u-component of wind (U$_{10M}$), and 10-meter v-component of wind (V$_{10M}$). Its original dimensions are 4384 $\times$ 6880 $\times$ 3, covering latitudes from 15° to 55° and longitudes from 75° to 135°. We used the data from May 2019 to October 2019 and July 2020 to August 2021 as training dataset, and from September 2021 as testing for data compression task. 

For the downscaling task, we used forecasts from the FuXi-2.0 model as low-resolution inputs. The FuXi-2.0 model is a cascaded machine learning weather forecasting model that provides 15-day global forecasts with a temporal resolution of 1 hour and a spatial resolution of 0.25 \textdegree \cite{chen2023fuxi,fuxi2.0}. In this study, we selected forecasts with lead times ranging from 1 to 24 hours initialized at 00 UTC and 12 UTC for our training set.

We selected 40 input variables at surface and pressure levels from the FuXi forecasts for downscaling task, guided by domain knowledge (see Table \ref{variables}). The dataset spans from May 2019 to October 2019 and July 2020 to August 2021, while the testing set includes data from November 2019 to June 2020 for testing the downscaling performance. The training set consists of 14,000 hours of data, whereas the testing set contains 6,346 hours, with no overlap between the two. Prior to inputting the data into the downscaling model, all input and output variables are normalized using z-score normalization.

\begin{table}[!ht]
    \centering
    \caption{The details of input variables.}\label{variables}
    \begin{tabular}{llll}
    \hline
        ~ & Variable & Full Name & Unit \\ \hline\hline
        Press Level & U(50, 200, 500, 700, 850, 925, 1000, hPa) & U-component of wind & m s$^{-1}$ \\ \hline
        ~ & V(50, 200, 500, 700, 850, 925, 1000, hPa) & V-component of wind & m s$^{-1}$ \\ \hline
        ~ & Z(50, 200, 500, 700, 850, 925, 1000, hPa) & Geopotential & m$^{2}$ s$^{-2}$ \\ \hline
        ~ & T(50, 200, 500, 700, 850, 925, 1000, hPa) & Temperature & K \\ \hline
        ~ & Q(50, 200, 500, 700, 850, 925, 1000, hPa) & Specific humidity & kg kg$^{-1}$ \\ \hline
        Surface & T$_{2M}$ & 2 meter temperature & K \\ \hline
        ~ & TP & Total precipitation & m h$^{-1}$ \\ \hline
        ~ & U$_{10M}$ & 10 meter U-component of wind & m s$^{-1}$ \\ \hline
        ~ & V$_{10M}$ & 10 meter V-component of wind & m s$^{-1}$ \\ \hline
        ~ & MSL & Mean sea level pressure & Pa \\ \hline
    \end{tabular}
\end{table}

\subsection{Overview of Data Compression Framework}

\textbf{Problem statement} We represent high-dimensional atmospheric data at a given time $t$ using a tensor X$^t$ with dimensions V $\times$ H $\times$  W, where H and W denote the number of latitude and longitude coordinates, respectively. For HRCLDAS, V=3, W=4384, H=6880. The indexing scheme X$^{t}_{v,h,w}$ indicates the value of variable $v$ at time t and latitude-longitude coordinates $(h, w)$. 

The first objective of this study is to compress X$^t$ to a lower-dimensional representation space Z$^t$ using an encoder with learnable parameters within the proposed VAE compression framework. The second goal is to utilize the compressed data for the downstream task of downscaling. Here, we select input Y${^{t}_v}$, which represents a variable with a lead time of $t$ from the FuXi forecast outputs. Our aim is to learn a mapping function $f$ that converts the low-resolution input state Y${_v}$ at timestamp $t$ to Z$^{t}$.

\begin{figure}[htb!]
\hspace{-1cm}
    \includegraphics[width=1.2\textwidth,height=0.6\textwidth]{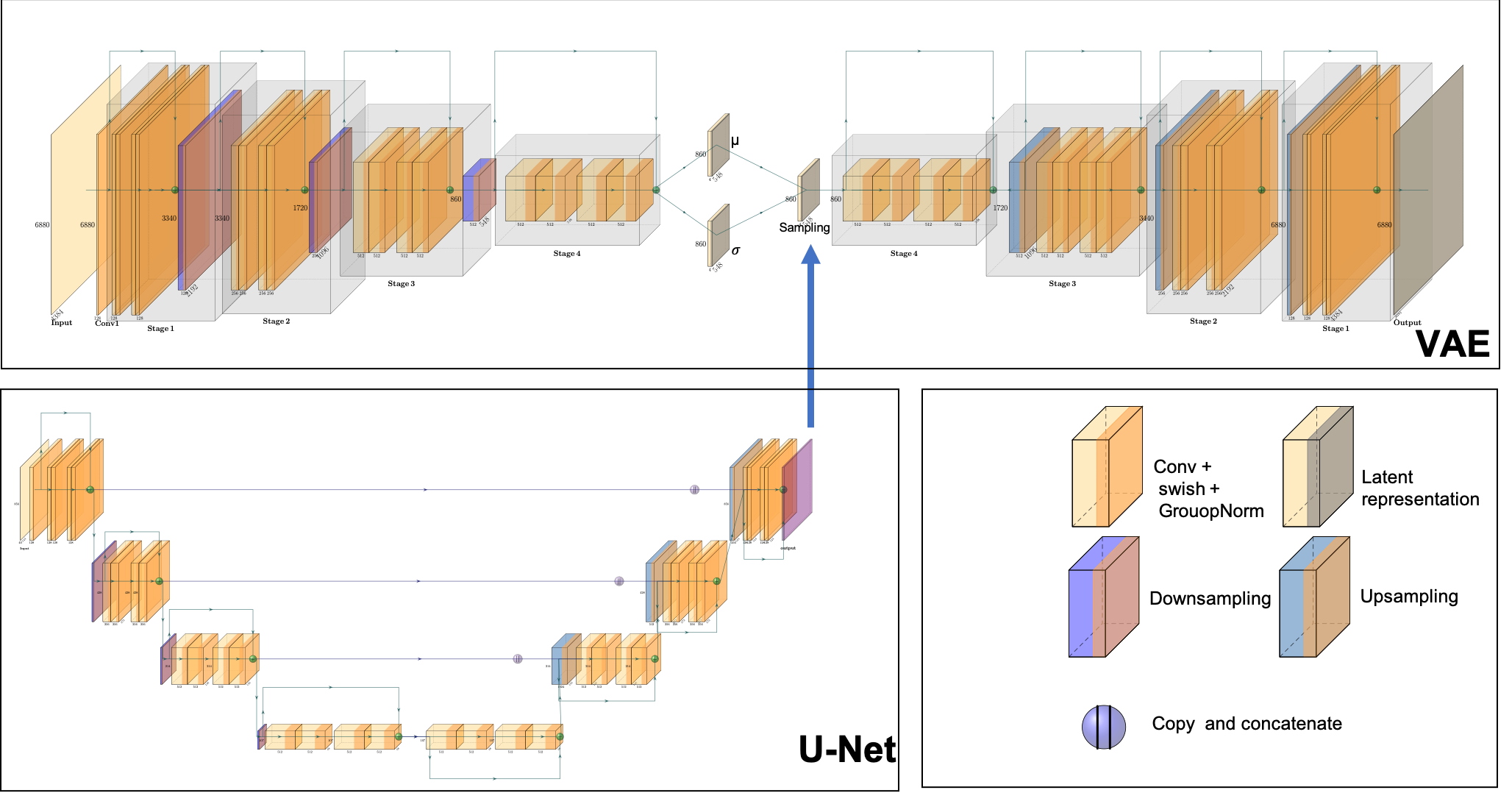}
    \caption{(Illustration of the proposed framework.}
    \label{fig:sr_earth}
\end{figure}

\textbf{VAE data compression framework}  Fig.\ref{fig:sr_earth} illustrates the overall compression framework based on the VAE neural network. The VAE consists of an encoder $E$ and a decoder $D$. 
The encoder $E$ compresses the information from the current timestamp into a low-dimensional latent vector Z. The decoder reconstructs the original high-resolution data $X'_{t,i}$ from latent vector Z through upsampling and convolution layers. To facilitate random sampling, $E$ outputs a mean $\mu$ and standard deviation $\sigma$, followed by a sampling function that generates the latent representation Z. 

Specifically, the encoder $E$ first transforms the input to a higher-dimensional feature space using a 1$\times$1$\times$C convolutional layer, with C (the number of channels) initially set to 128. This is followed by four stages, where the first three each contain two ResNet blocks and a downsampling block. Each ResNet block utilizes two convolution layers with a kernel size of 3, followed by a Swish activation function \citep{ramachandran2017searching} and group normalization layers. Residual connections link the input and output, mitigating the risk of vanishing gradients. 

The first convolutional layer of each stage doubles the feature map size, while the second convolutional layer maintains the same size. At the end of each stage, a downsampling block is applied, consisting of a convolutional layer with a kernel size of 3 and a stride of 2 to reduce the volume by half, replacing the traditional pooling operation. The fourth stage differs in that it does not include a downsampling layer. Instead, it uses two additional ResNet blocks to refine the compressed features. Finally, a convolutional layer with a feature map size of 4 is used to generate the mean $\mu$ and var $\sigma$,both with dimensions 548$\times$860$\times$4. 

The decoding process is a symmetric process with the encoding process. The decoder initially uses a convolutional layer to convert the channel dimension from 4 to 512. Bilinear interpolation method is used for upsampling at the end of each stage. Two ResNet blocks in the decoder further refine the features. The four decoding stages process the features sequentially, yielding feature map sizes of 548$\times$860$\times$4, 548$\times$860$\times$512, 548$\times$860$\times$512, 1096$\times$1720$\times$256, and 2192$\times$3440$\times$128, respectively. Finally, a convolutional layer converts the channel dimension to the output dimension, resulting in an output size of 4384$\times$6880$\times$3.

The loss component of the VAE consists of the reconstruction loss and the Kullback–Leibler divergence $D_{kl}$. We utilized the charbonnie loss \citep{lai2018fast} as reconstruction loss function as described in Eq.~\ref{charbonnie} to better handle the outliers situation. In addition, the Kullback–Leibler divergence $D_{kl}$ regularizes the posterior distribution of the latent space $P = N(\mu,\sigma^2)$ to match the prior distribution, a standard Gaussian distribution  $Q=N(0,1)$. Those parameters are optimized with the help of the re-parametrization trick \citep{kingma2016improved}.


\begin{equation}\label{charbonnie}
    \mathcal{L}(X, X') = \sqrt{||X -X'||^2 + \epsilon ^2}
\end{equation}

\begin{equation}
    KL(P||Q) = \sum_{x}P(X)log(\frac{P(X)}{Q(X)})
\end{equation}\label{eq:kl}

\textbf{Downscaling Model:} For the downstream task, we selected downscaling as a case study to evaluate the usability of the compact data generated by our proposed VAE framework. The primary objective of the downscaling model is to enhance the output resolution of FuXi forecasts. Once the VAE model is trained, the HRCLDAS data are encoded by the VAE encoder to extract latent representations, which serve as ground truth for the downscaling model. We employed U-Net for this task\citep{ronneberger2015u}. The original FuXi forecast input has dimensions of 176$\times$276 with a spatial resolution of 25 km. For downscaling, we interpolate the data to a resolution of 550$\times$3862$\times$3 with a spatial resolution of 8 km. 

U-Net utilizes a similar encoder-decoder architecture as the VAE. It comprises four stages, each containing two ResNet blocks in both the encoder and decoder.  The decoder $D_{unet}$, which serves as an expansive path, employs upsampling layers to project the encoder output to the resolution of the compact data. The decoding part of U-Net is connected to the corresponding encoder part via skip connections. A convolution operation is applied to the concatenated output of the up-sampled output and its encoder counterpart, as illustrated in Fig.~\ref{fig:sr_earth}. Eventually, the U-Net trained on the original HRCLDAS data has 118 million trainable parameters, while the U-Net trained on compressed data has 94.8 million parameters.

\subsection{Implementation details}

We first trained a VAE for data compression. The model was trained with a batch size of 8 and optimized using the Adam optimizer with a learning rate of 1.6e-5. Given the high resolution of the original HRCLDAS data, we split it into 1000$\times$1000 patches with overlaps in the latitude direction, resulting in 35 patches per dataset. To reduce computational costs and accelerate convergence, we further divided these 1000$\times$1000 patches into smaller 256$\times$256 patches. We pre-trained the VAE model on these smaller patches for the first 10 epochs and subsequently fine-tuned it using the larger patches for an additional 5 epochs. Unlike natural images, where channels are highly correlated, weather variables exhibit weak inter-variable relationships. Therefore, we trained separate VAE models for T${_{2m}}$, and U$_{10M}$, V$_{10M}$, respectively. 

For training the downscaling model, we used the compressed data Z$^{t}$ with a resolution of 544$\times$856 generated from the trained VAE encoder as ground truth. We trained the U-Net model with a batch size of 16 for 50 epochs. The model was optimized using the Adam optimizer with a learning rate of 3.2e-5. All the models were implemented using the PyTorch framework and trained on 8 NVIDIA A100 GPUs.

\subsection{Experiment setup}

Four experiments were conducted to assess the impact of different modeling strategies on data reconstruction performance within the proposed VAE framework. Additionally, three experiments were performed to evaluate the model's downscaling performance using the compact data generated by the VAE. 

The four experiments within the VAE framework are detailed in Table ~\ref{tab:vae_training}. The baseline model (resize) employs a straightforward downsampling technique to reduce dimensions and an upsampling method for data reconstruction. To achieve improved reconstruction results, we trained the two VAE models using two different loss functions: $L1$ loss and Charbonnier loss, on 256 $\times$ 256 patches. Since the target variables do not exhibit linear correlations with one another, we trained separate models (VAE single variable) for each target variable, rather than treating them collectively, as is common with natural images. Finally, to obtain more accurate results, we fine-tuned the model trained on smaller patches using the full resolution data (VAE fine-tune). 

For the downscaling task, we used bilinear interpolation (Inter) as the baseline for comparsion. We trained the U-Net model on both the original HRCLDAS data (No-VAE) and the compact data generated by the encoder from VAE (fine-tune) model, respectively, to access the impact of the data compression method on downstream task performance.


\begin{table}[h]
\caption{The setting  of different  data  compression methods}\label{tab:vae_training}%
\begin{tabular}{@{}lllll@{}}
\toprule
method & $L1$ loss  & charbonnie loss 3 & single variable & fine tune\\
\midrule
Resize &  &  &  &  \\
VAE ($L1$)& $\checkmark$  &   &   &  \\
VAE(charbonnie loss) &  & $\checkmark$  \\
VAE (single variable) &  & $\checkmark$ & $\checkmark$ &  \\
VAE(fine-tune) &  & $\checkmark$ & $\checkmark$ & $\checkmark$  \\
\botrule
\end{tabular}
\end{table}

\subsection{Evaluation metrics}

To evaluate the performance of the reconstruction and its downscaling capabilities, we use several metrics commonly applied in the weather and climate domain \citep{rasp2024weatherbench,zhong2024investigating,gong2022temperature}. Specifically, we calculate the mean square error (MSE), root mean square error (RMSE), and power spectrum.

The MSE measures the difference between the ground truth $X_{w,h}$ and the reconstructed or downscaled results $X'_{w,h}$, where the grid corresponds to the cell center positions $w$ and $h$ in the zonal and meridional directions, respectively. RMSE, on the other hand, represents the average difference between the ground truth and the reconstructed data generated by the model. Both MSE and RMSE are negatively oriented, with an ideal value of 0 indicating perfect reconstruction.

\begin{equation}\label{eq:MSE}
    MSE(v) = \frac{1}{N} \sum_{w=1}^{W} \sum_{h=1}^{H} \left[X'_{v,h,w}-X_{v,h,w}\right]^{2}.
\end{equation}

\begin{equation}\label{equ:rsme}
RMSE(v) = \sqrt{\frac{1}{N} \sum_{w=1}^{W} \sum_{h=1}^{H} (X'_{v,h,w}-X_{v,h,w})^{2}}
\end{equation}

In addition, to evaluate the local-scale variability and determine the information preserved or lost by the VAE framework for the downstream tasks, we conduct power spectrum analysis following the methodology described by \citet{rasp2024weatherbench}. The power spectrum is calculated along lines of constant latitude as a function of wavenumber (unitless), frequency (km$-$1), and wavelength (km). The Discrete Fourier Transform (DFT), denoted as $F_{K}$ is computed using the following Eq.\ref{df}:

\begin{equation}
    F(k) = \frac{1}{L} \sum_{l=0}^{L-1} f_{l}e^\frac{{-i2 \pi kl}}{L}
\end{equation}\label{df}

Since our evaluation focuses on the regional rather than global scales, we do not consider the circumference in this study. The power spectrum for constant latitude is obtained using the following equation:

\begin{equation}
Sk = 2|F_{k}|^2, k = 1,2,..., L/2
\end{equation}\label{equ:ps} 

Additionally, we approximate the average zonal power spectrum using:
\begin{equation}\label{equ:zps} 
\int_{0}^{l} |f{l} |dl \approx \sum_{k=0}^{L/2} S_k
\end{equation}

In addition to the meteorological evaluation, we incorporate the structural similarity index (SSIM) \citep{wang2004image}, which was initially developed for computer vision to quantify and compare spatial variability between downscaled fields and ground truth data. This metric has recently been applied in meteorological research \citep{zhong2024investigating,gong2022temperature}.

\begin{equation}
SSIM(y,\hat{y}) = \frac{(2\mu_{\hat{y}}\mu_{y} + C_1) + (2 \sigma_{\hat{y}y} + C_2)}{(\mu_{\hat{y}^2} + \mu_{y^2}+C_1)(\sigma_{\hat{y}^2)} + \sigma_{y^2}+C_2)}
\end{equation}

\section{Experiments results}\label{sec3}

\subsection{Representation performance}\label{subsec2}


\begin{table}[!ht]
    \centering
    \caption{The performance in terms of RMSE for different data compression models, where U$_{10M}$, V$_{10M}$, and T$_{2M}$ represent 10 meter U-component of wind, 10 meter V-component of wind, and 2 meter temperature.}\label{tab:vae_perform}    
    \begin{tabular}{llll}
    \hline
     RMSE&  U$_{10M}$ & V$_{10M}$ & T$_{2M}$ \\ \hline\hline
        Resize & 0.0771 & 0.0783 & 0.7109 \\ \hline
        VAE ($L1$)& 0.108 & 0.100 & 0.752 \\ \hline
        VAE(charbonnie loss) & 0.104 & 0.090 & 0.578 \\ \hline
        VAE (single variable) & 0.0428 & 0.029 & 0.353 \\ \hline
        VAE(fine-tune) & \textbf{0.0181} & \textbf{0.0139}& \textbf{0.124}\\ \hline
    \end{tabular}
\end{table}

Table \ref{tab:vae_perform} presents performance results for various data compression strategies, including simple resize, VAE with $L$1 loss, VAE with Charbonnier loss, VAEs trained on single variables, and the VAE fine-tuning strategy. Among these methods, the VAE fine-tuning strategy demonstrates superior performance in reconstructing U$_{10M}$, V$_{10M}$, and T$_{2M}$. Additionally, we observed that simple VAE with $L1$ and Charbonnie loss struggles to effectively represent the temperature and wind fields, leading to a high RMSE. Notably, training on a single variable is consistently superior to training on multiple variables. Furthermore, a comparison between the VAE trained from scratch (single variable) and the VAE fine-tuned from pretraining with small patches to full resolution indicates that the pretraining method significantly enhances the representation performance. The VAE fine-tuning approach achieves RMSE values of 0.0181 $K$, 0.0139 $K$, and 0.124 $K$ for the three variables respectively, which is more than 4 times better than the linear interpretation method. 

\begin{figure*}[ht!]
    \begin{subfigure}[t]{0.3\textwidth}
        \includegraphics[width=\linewidth]{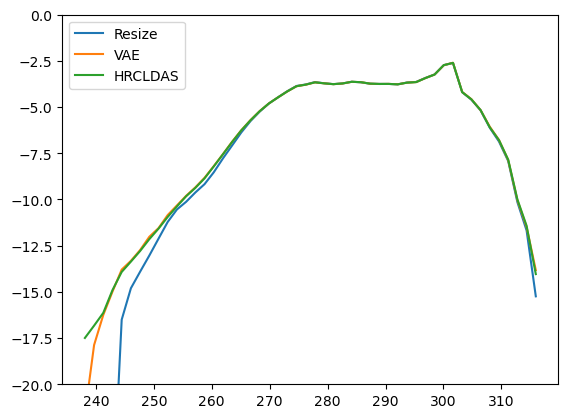}
         \caption{T$_{2M}$}
    \end{subfigure}
     \hspace{-0.1cm}
    \begin{subfigure}[t]{0.3\textwidth}
        \includegraphics[width=\linewidth]{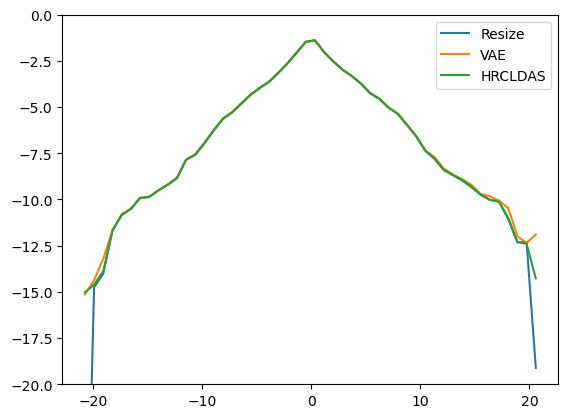}\caption{U$_{10M}$}
    \end{subfigure} 
    \hspace{-0.1cm}
    \begin{subfigure}[t]{0.3\textwidth}
        \includegraphics[width=\linewidth]{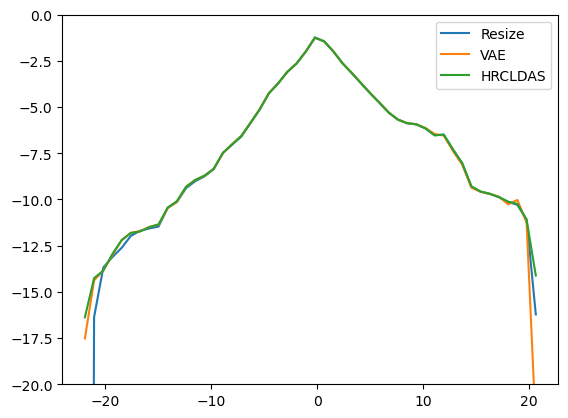}
       \caption{V$_{10M}$}
    \end{subfigure}

    \caption{Displaying of log(density) plot for (a) T$_{2M}$, (b) U$_{10M}$, and (c) V$_{10M}$ by the VAE method comparing to the resize method and HRCLDAS data.}\label{fig:density}
\end{figure*}

To further understand the VAE's performance in data compression, particularly for extreme values, we examine the differences in density distributions among the ground truth, VAE, and a simple resize baseline model, as illustrated in Fig.~\ref{fig:density}. For T$_{2M}$ (Fig.~\ref{fig:density}a), the VAE-reconstructed data closely aligns with the frequency of positive temperature values in the ground truth HRCLDAS data. In contrast, the resize method results in a slight decrease in the frequency of positive temperatures. For negative temperatures, the VAE model exhibits a lower frequency compared to the original HRCLDAS data. Regarding U$_{10M}$ and V$_{10M}$ (Fig.~\ref{fig:density}b and Fig. ~\ref{fig:density}c), the VAE effectively preserves a higher frequency of both extreme high and low values.

\begin{figure*}[ht]
    \hspace{1 cm}
    \begin{subfigure}[t]{0.3\textwidth}
        \includegraphics[width=\linewidth]{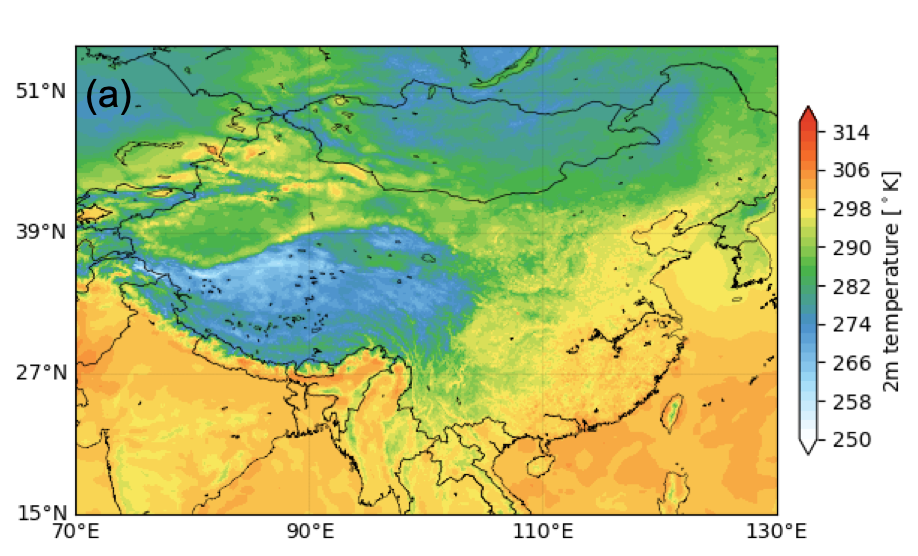}
    \end{subfigure}
    \hspace{-1cm}
    \begin{subfigure}[t]{0.3\textwidth}
        \includegraphics[width=\linewidth]{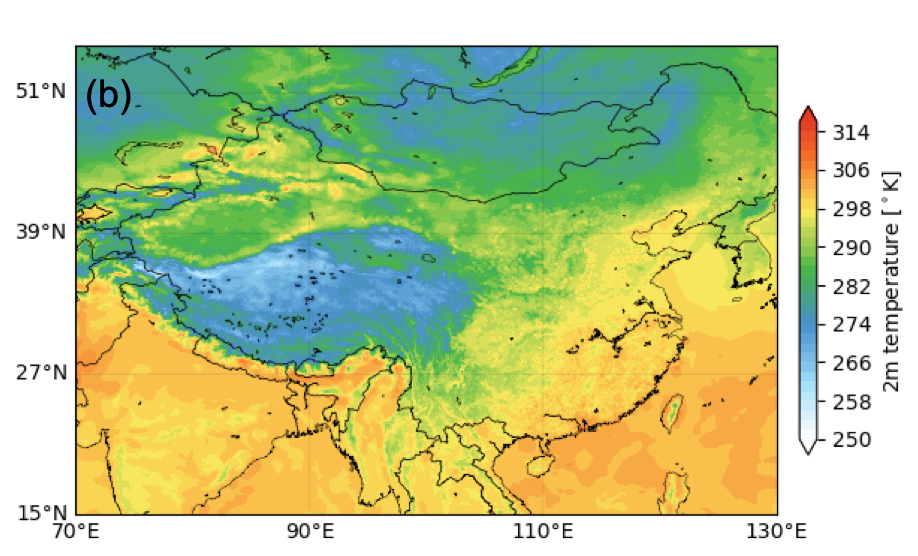}
    \end{subfigure}
     \hspace{-0.3cm}
    \begin{subfigure}[t]{0.3\textwidth}
        \includegraphics[width=\linewidth]{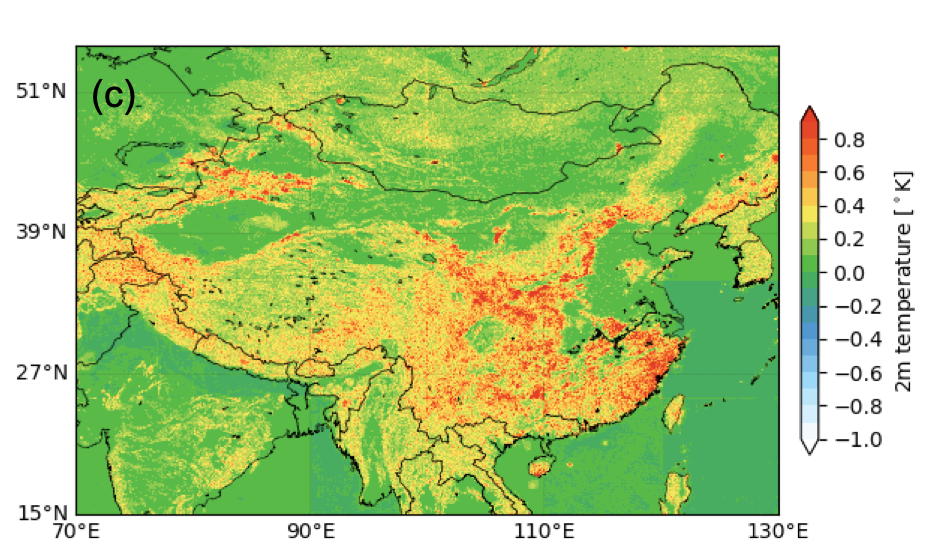}
    \end{subfigure}
    
    \hspace{1 cm}
    \begin{subfigure}[t]{0.3\textwidth}
        \includegraphics[width=\linewidth]{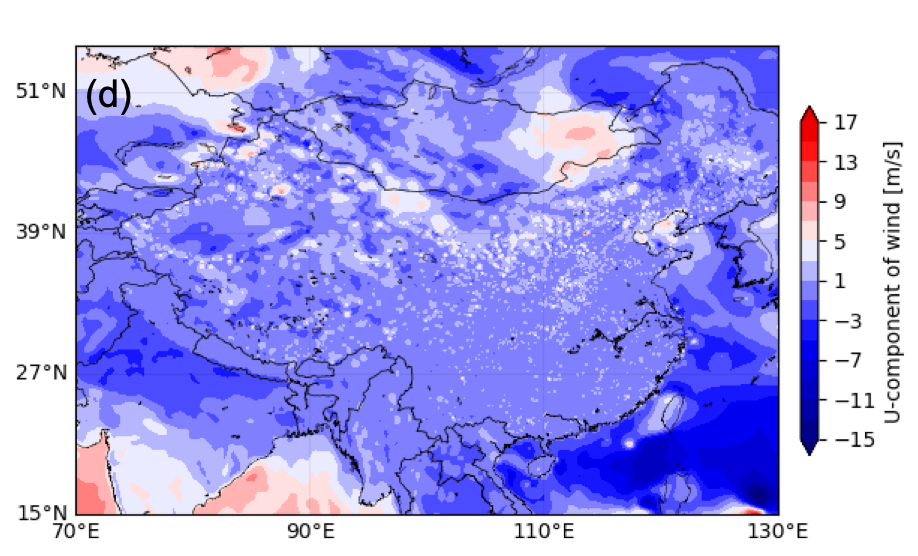}
    \end{subfigure}
    \hspace{-1cm}
    \begin{subfigure}[t]{0.3\textwidth}
        \includegraphics[width=\linewidth]{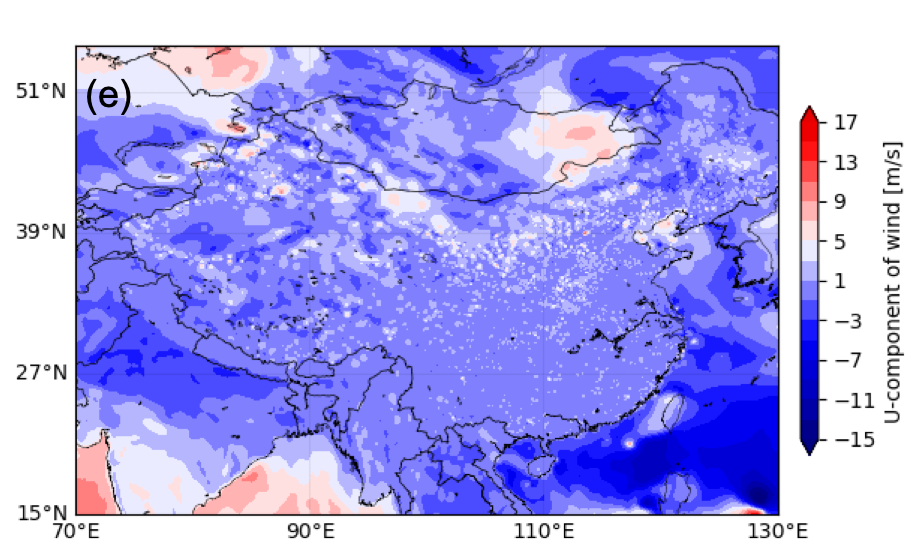}
    \end{subfigure}
     \hspace{-0.3cm}
    \begin{subfigure}[t]{0.3\textwidth}
        \includegraphics[width=\linewidth]{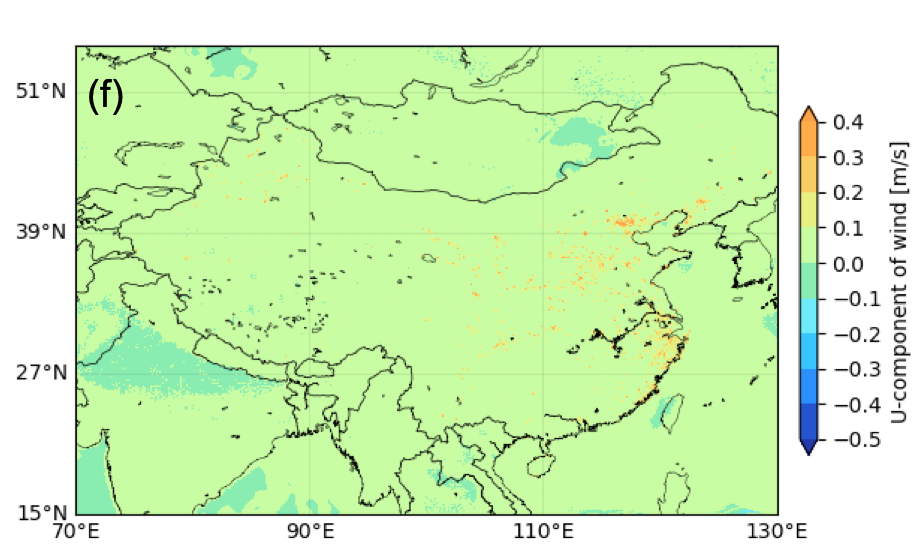}
    \end{subfigure}
    
    \hspace{1 cm}
    \begin{subfigure}[t]{0.3\textwidth}
        \includegraphics[width=\linewidth]{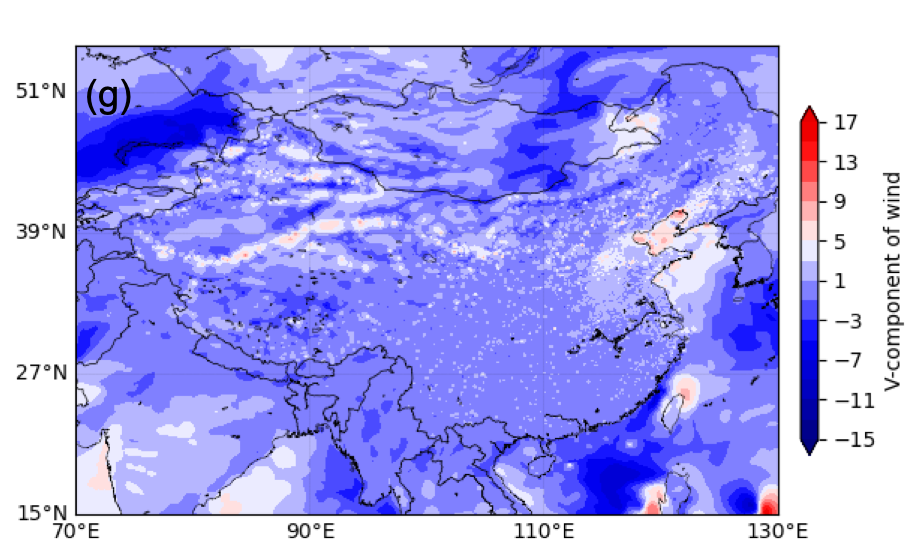}
    \end{subfigure}
    \hspace{-1cm}
    \begin{subfigure}[t]{0.3\textwidth}
        \includegraphics[width=\linewidth]{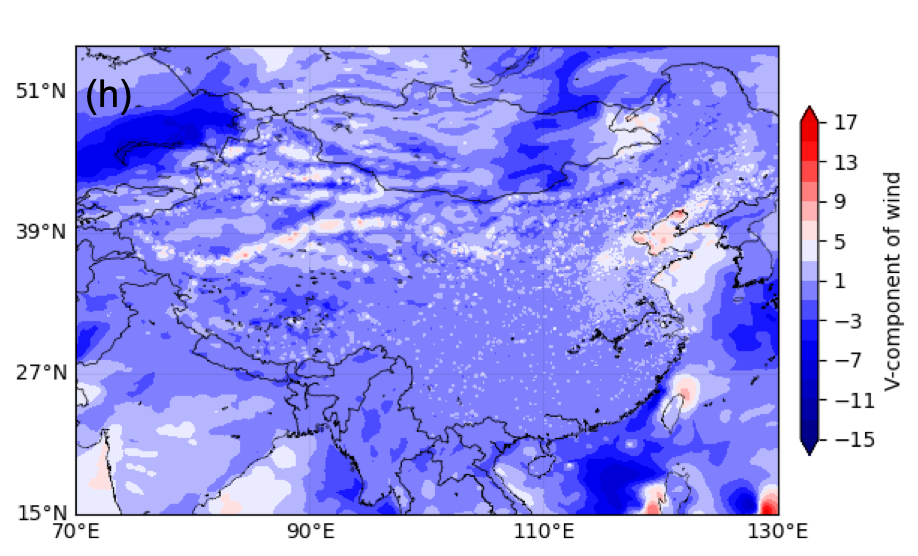}
    \end{subfigure}
     \hspace{-0.3cm}
    \begin{subfigure}[t]{0.3\textwidth}
        \includegraphics[width=\linewidth]{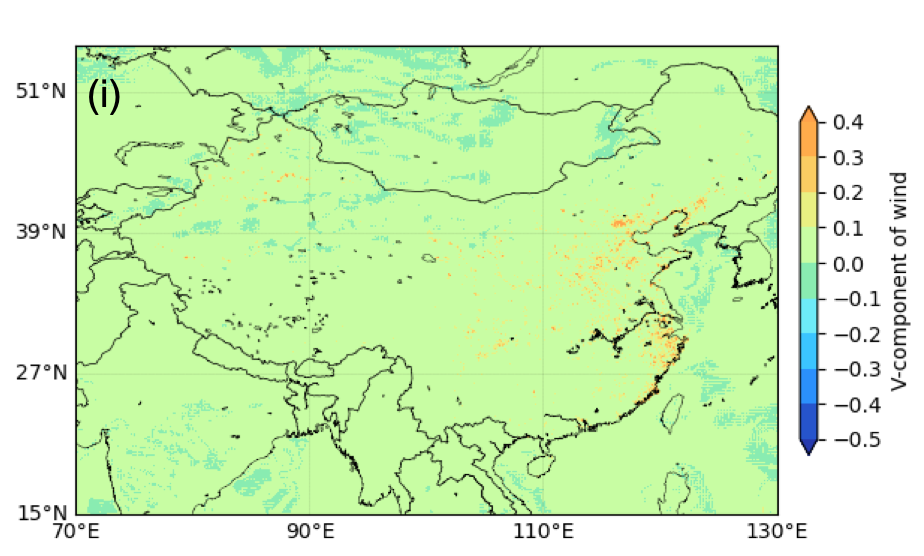}
    \end{subfigure}
    \caption{Example reconstructed fields by VAE (fine-tune) for T$_{2M}$, U$_{10M}$ and V$_{10M}$ respectively on 9\ts{th} September 2021, 00 UTC. (a) T$_{2M}$ from HRCLDAS, (b) the reconstructed T$_{2M}$ field by VAEs (fine-tune), (c) differences between HRCLDA T$_{2M}$ and reconstructed T$_{2M}$, (d) U$_{10M}$ from HRCLDAS, (e) the reconstructed U$_{10M}$ field by VAEs (fine-tune), (f) differences between HRCLDA U$_{10M}$ and reconstructed U$_{10M}$, (g) V$_{10M}$ from HRCLDAS, (h) the reconstructed V$_{10M}$ field by VAEs (fine-tune), (g) differences between HRCLDA V$_{10M}$ and reconstructed V$_{10M}$.}
\end{figure*}

\subsection{Downscaling performance}\label{subsec2}

\begin{figure*}[ht!]
    \begin{subfigure}[t]{0.3\textwidth}
        \includegraphics[width=\linewidth]{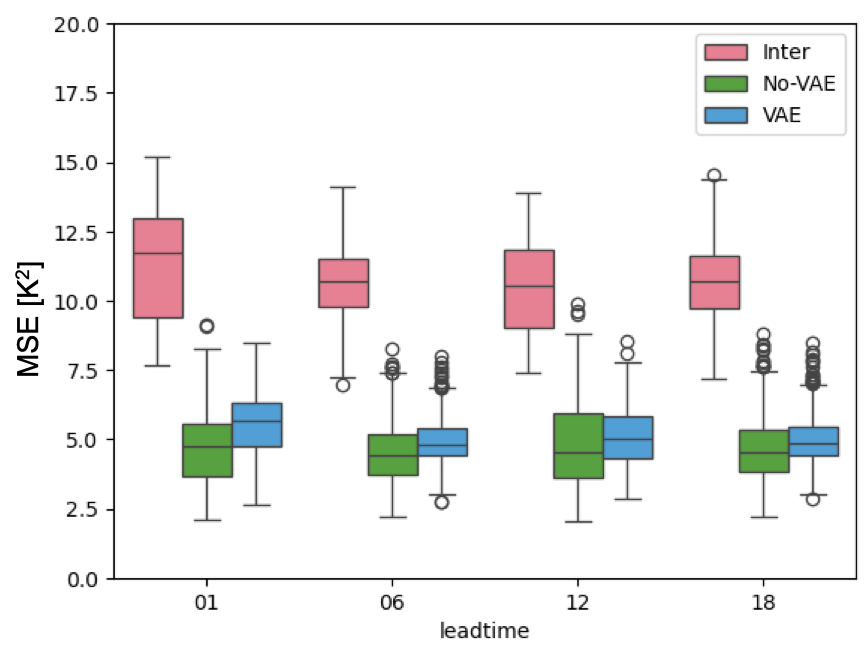}\caption{T$_{2M}$}
    \end{subfigure} 
    \hspace{-0.1cm}
    \begin{subfigure}[t]{0.3\textwidth}
        \includegraphics[width=\linewidth]{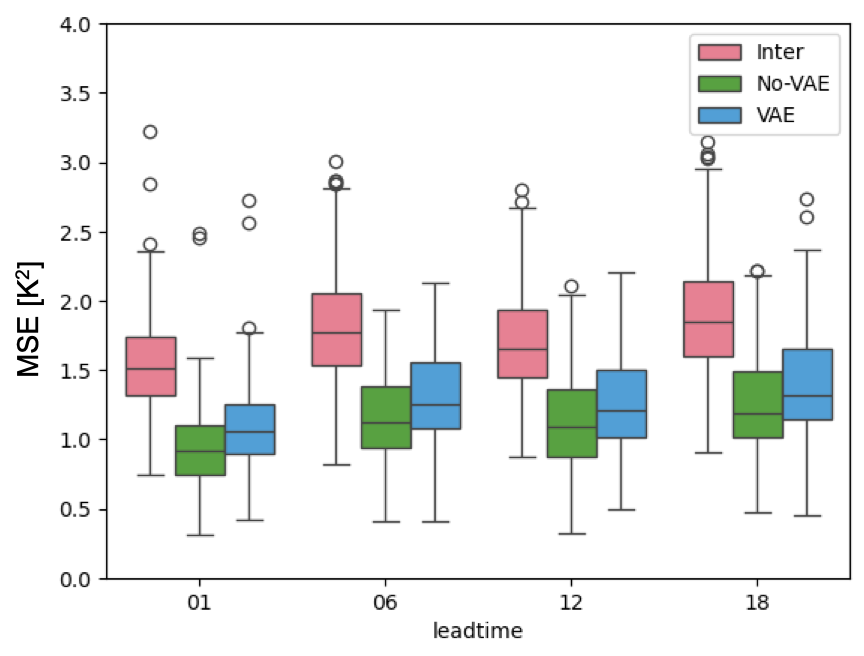}
        \caption{U$_{10M}$}
    \end{subfigure}
     \hspace{-0.1cm}
    \begin{subfigure}[t]{0.3\textwidth}
        \includegraphics[width=\linewidth]{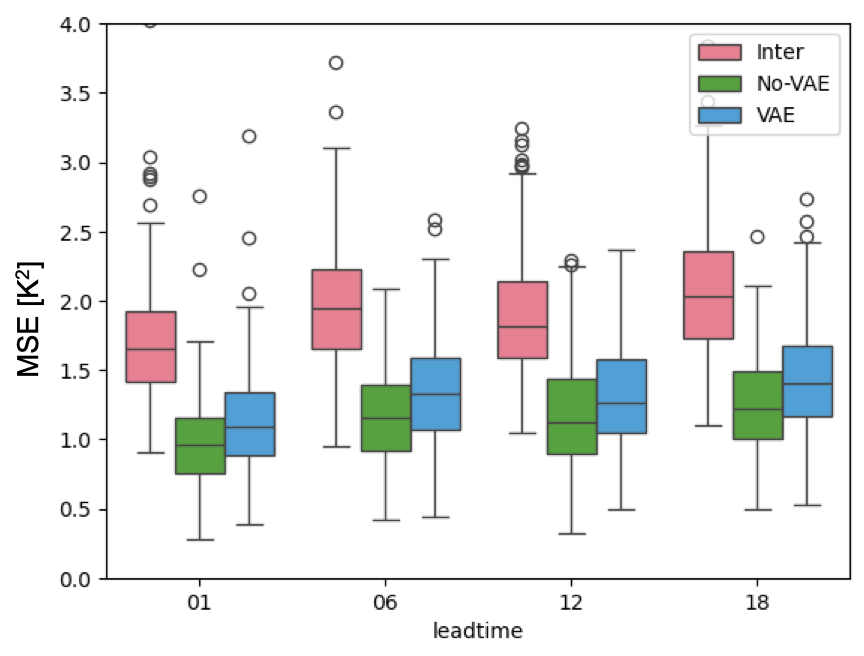}
        \caption{V$_{10M}$}
    \end{subfigure}
    \caption{Box-and-whisker plot for comparision in terms of MSE between baseline and machine learning methods for (a) T$_{2M}$, (b) U$_{10M}$, and (c) V$_{10M}$ with lead time of 1h to 18h. The solid horizon indicates the minimum and maximum MSE, excluding outliers; the box bounds the interquartile range from the 25th to 75th percentiles, with the 50th percentile. The red color indicates the baseline model "resize"; the green color indicates the U-Net trained on original HRCLDAS data; the blue color indicates the U-Net trained on compact HRCLDAS data generated by VAE.}\label{fig:mse}
\end{figure*}

\begin{figure*}[ht!]
    \begin{subfigure}[t]{0.3\textwidth}
        \includegraphics[width=\linewidth]{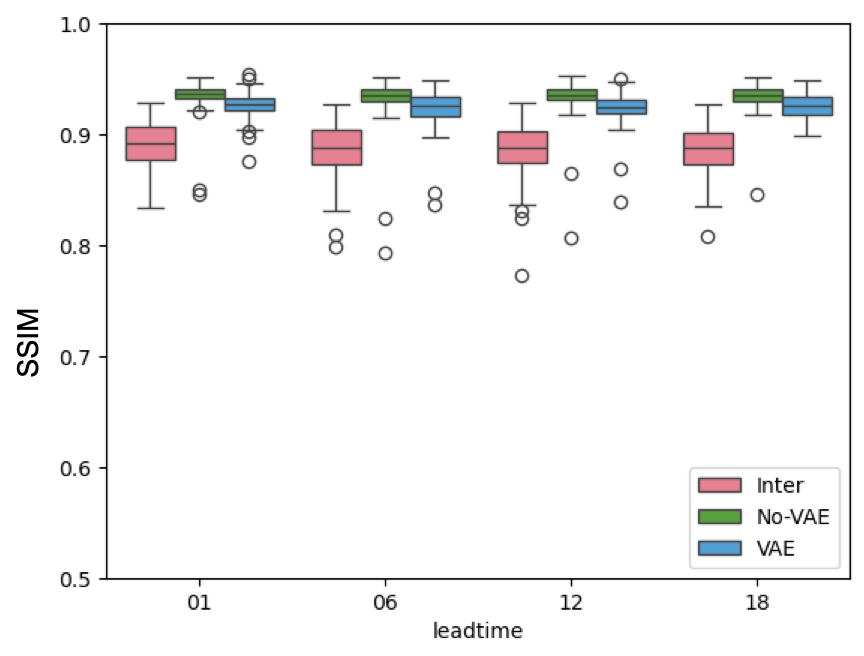}\caption{T$_{2M}$}
    \end{subfigure} 
    \hspace{-0.1cm}
    \begin{subfigure}[t]{0.3\textwidth}
        \includegraphics[width=\linewidth]{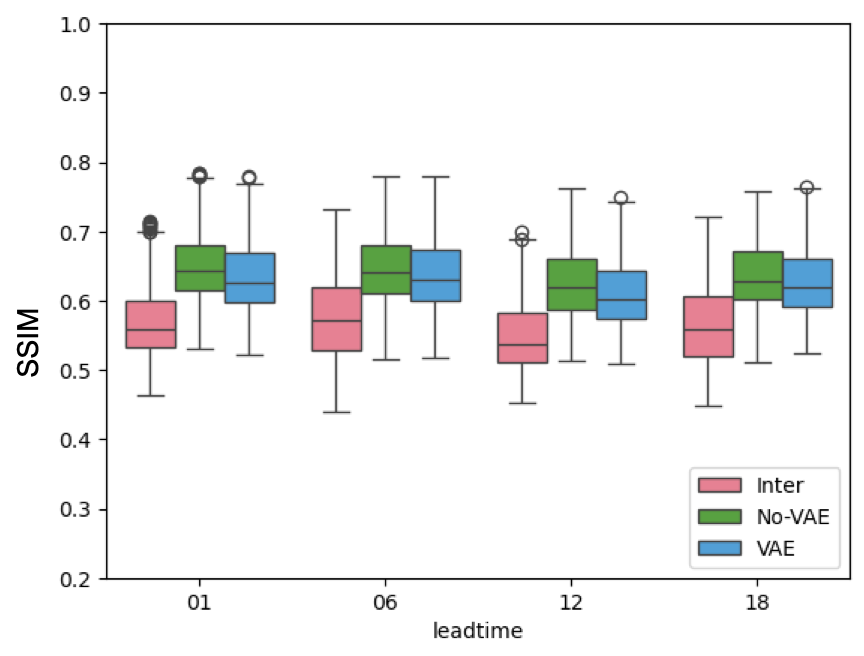}
       \caption{U$_{10M}$}
    \end{subfigure}
     \hspace{-0.1cm}
    \begin{subfigure}[t]{0.3\textwidth}
        \includegraphics[width=\linewidth]{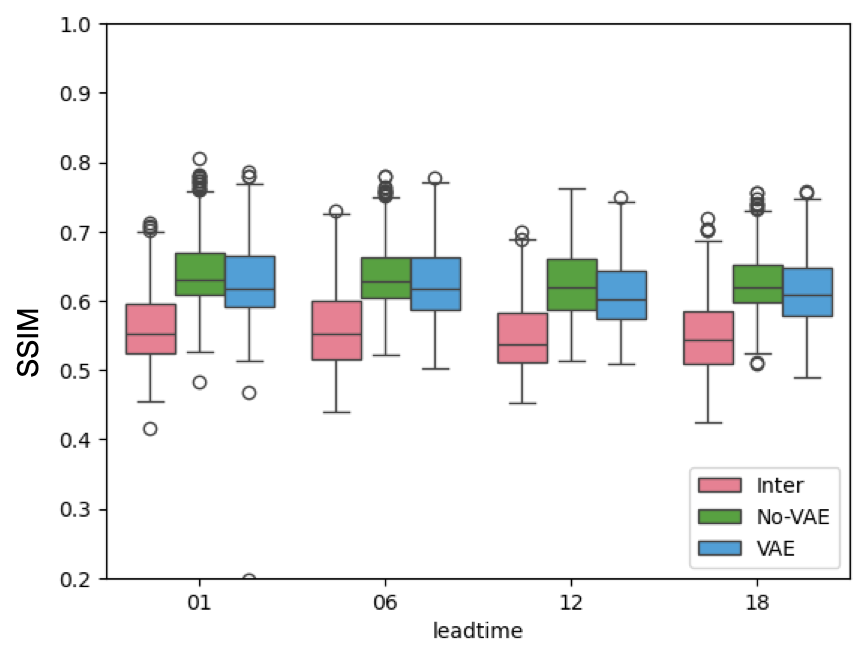}
         \caption{V$_{10M}$}
    \end{subfigure}
    \caption{The Box-and-whisker plot for comparsion in terms of SSIM between baseline and machine learning methods for (a) T$_{2M}$, (b) U$_{10M}$, and (c) V$_{10M}$ with a lead time of 1~h to 18~h. The solid horizon indicates the minimum and maximum MSE, excluding outliers; the box bounds the interquartile range from the 25th to 75th percentiles, with the 50th percentile. The red color indicates the baseline model "resize"; the green color indicates the U-Net trained on original HRCLDAS data; the blue color indicates the U-Net trained on compact HRCLDAS data generated by VAE.}\label{fig:ssim}
\end{figure*}

In our study, we selected the fine-tuned VAE (fine-tune) as the data compression method, subsequently applying the compressed data produced by VAE (fine-tune) for our downstream task - downscaling. Fig.~ \ref{fig:mse} and Fig ~\ref{fig:ssim} demonstrate the performance of the UNet model trained on compressed HCLDAS data (VAE) and original data (No-VAE) in terms of MSE (Fig. \ref{fig:mse}) and SSIM( Fig.\ref{fig:ssim}) respectively, compared to the baseline model - interpolation method, across lead times from 1 to 18 hours. The results indicate that the deep learning-based model significantly outperforms the simple interpolation method (Inter). Specifically, the mean square error values for linear interpolation, No-VAE, and VAE are 11.58 $K^2$, 4.965,79$K^2$, and 5,79$K^2$, respectively, demonstrating that the No-VAE and VAE outperform linear interpolation by approximately 53\%, 60\% for temperature in terms of MSE.

We also employed the SSIM, commonly used in downscaling tasks to access the perceptual similarity between images. SSIM quantifies and compares mean and global spatial variability in the downscaled fields against the ground truth and also accounts for covariances. The SSIM comparisons for all models are presented in Fig \ref{fig:ssim}, showing that both VAE and No-VAE consistently outperform the interpolation method across all lead times and variables. 

\begin{figure*}[ht!]
    \begin{subfigure}[t]{0.3\textwidth}
        \includegraphics[width=\linewidth]{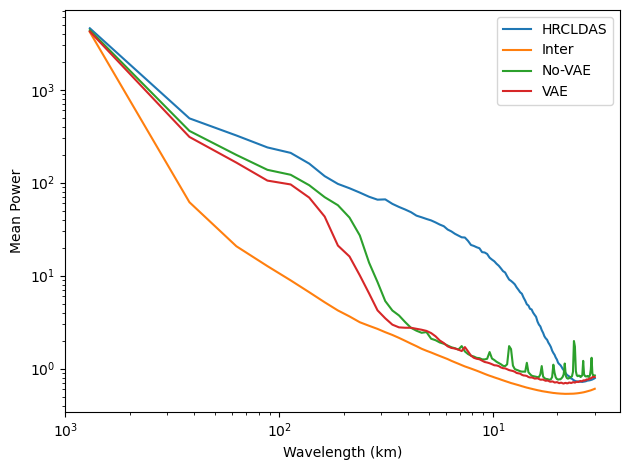}\caption{T$_{2M}$}
    \end{subfigure} 
    \hspace{-0.1cm}
    \begin{subfigure}[t]{0.3\textwidth}
        \includegraphics[width=\linewidth]{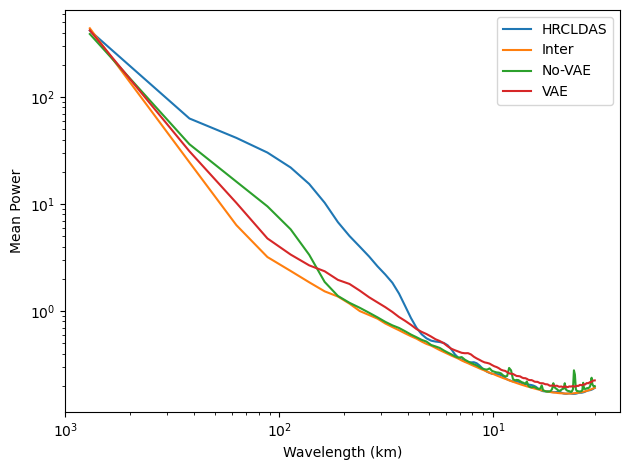}
       \caption{U$_{10M}$}
    \end{subfigure}
     \hspace{-0.1cm}
    \begin{subfigure}[t]{0.3\textwidth}
        \includegraphics[width=\linewidth]{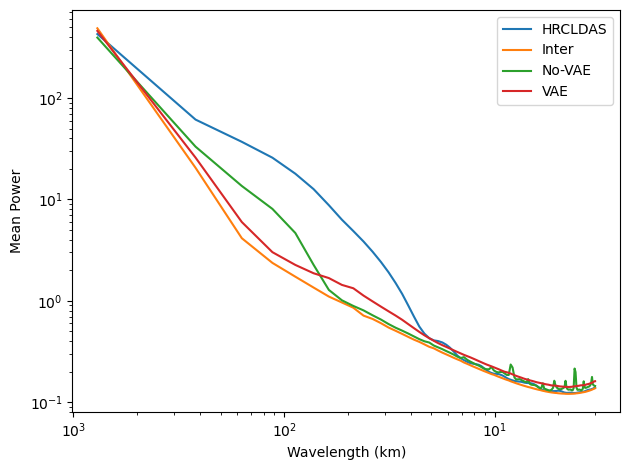}
         \caption{V$_{10M}$}
    \end{subfigure}
    \caption{Plot displaying of the power spectrum for (a) T$_{2M}$, (b) U$_{10M}$, and (c) V$_{10M}$ with a lead time of 1~h by the VAE method comparing to the interpolation, Non-VAE, and HRCLDAS ground truth data. The x-axis indicates the number of wavelength (~km) and y-axis is the number average power spectrum.}\label{fig:ps}
    
\end{figure*}

\begin{figure*}[ht!]

    \begin{subfigure}[t]{0.27\textwidth}
        \includegraphics[width=\linewidth]{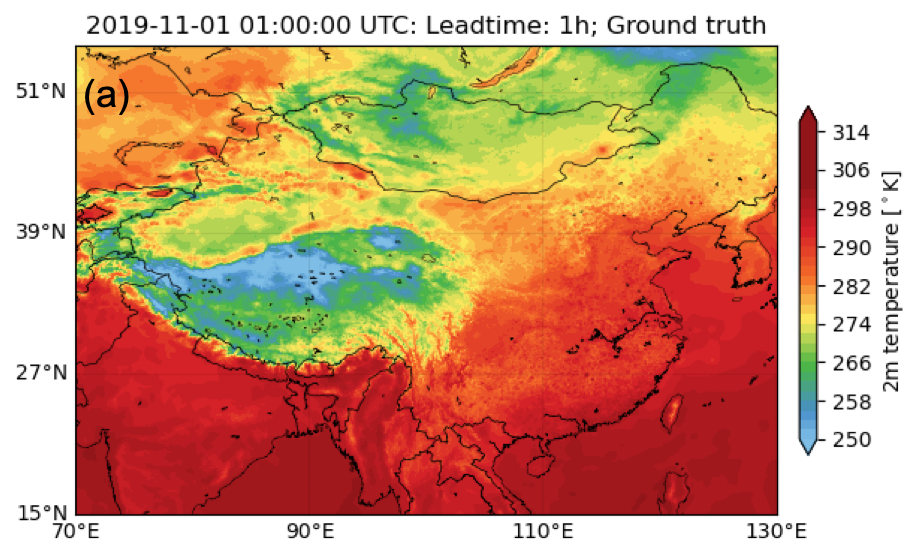}
    \end{subfigure}
    \hspace{-0.8cm}
    \begin{subfigure}[t]{0.27\textwidth}
        \includegraphics[width=\linewidth]{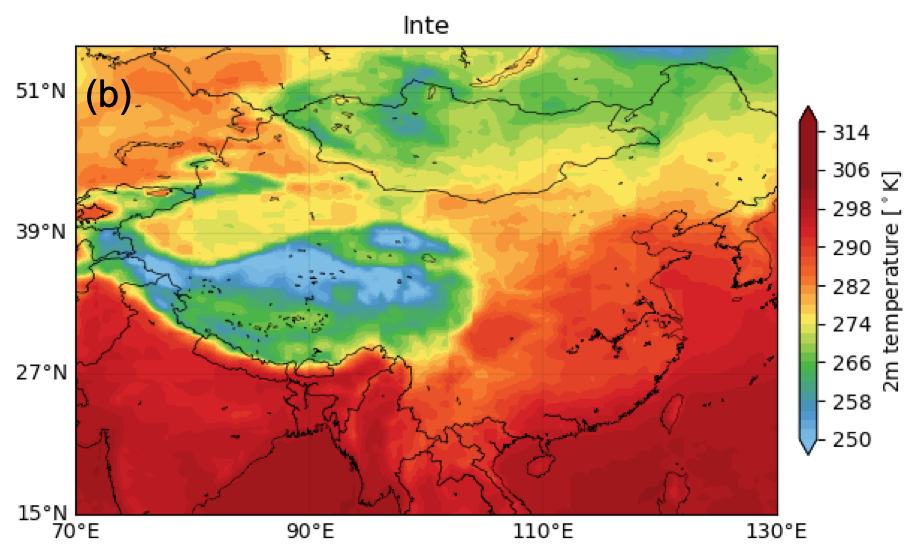}
    \end{subfigure}
     \hspace{-0.8cm}
    \begin{subfigure}[t]{0.27\textwidth}
        \includegraphics[width=\linewidth]{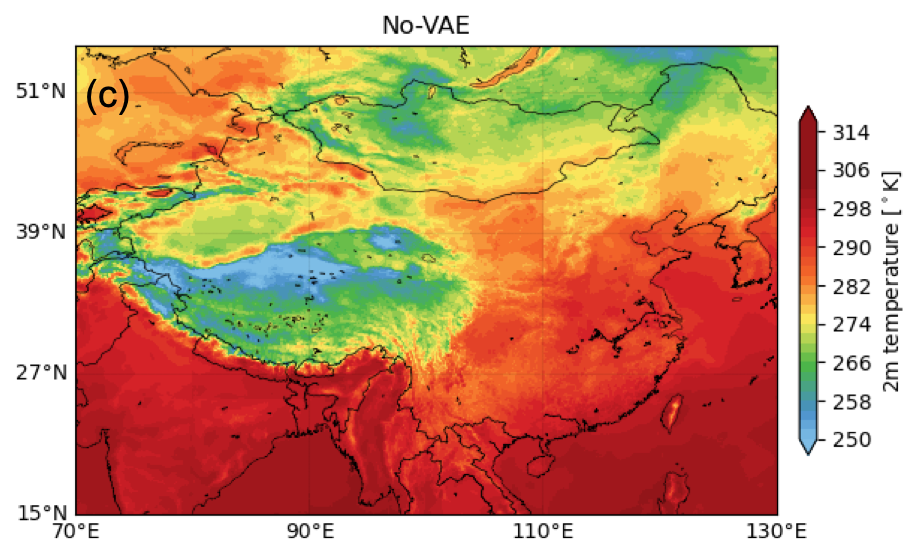}
    \end{subfigure}
    \hspace{-0.8cm}
    \begin{subfigure}[t]{0.27\textwidth}
        \includegraphics[width=\linewidth]{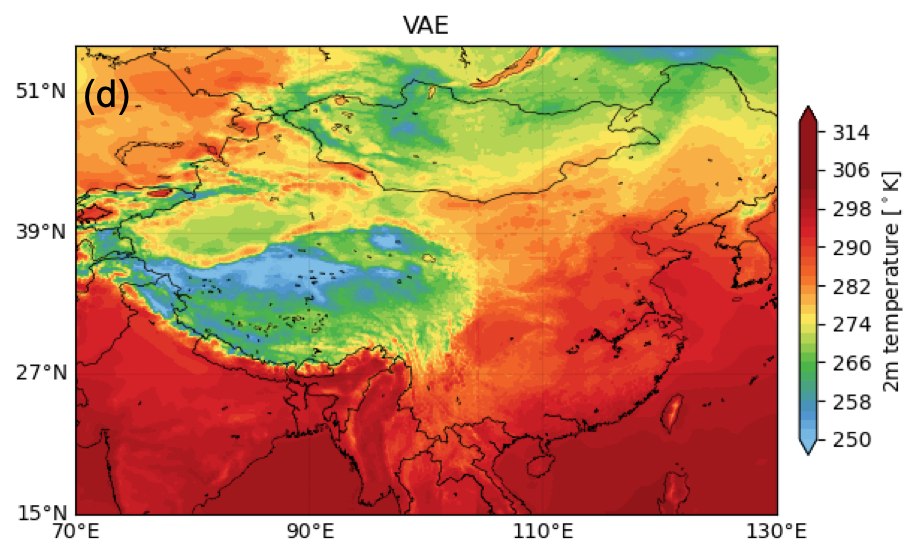}
    \end{subfigure}

    \begin{subfigure}[t]{0.27\textwidth}
        \includegraphics[width=\linewidth]{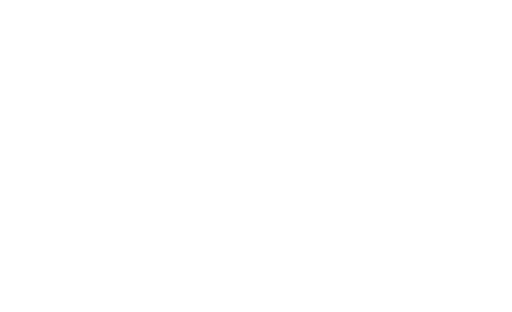}
    \end{subfigure}
     \hspace{-0.8cm}
    \begin{subfigure}[t]{0.27\textwidth}
        \includegraphics[width=\linewidth]{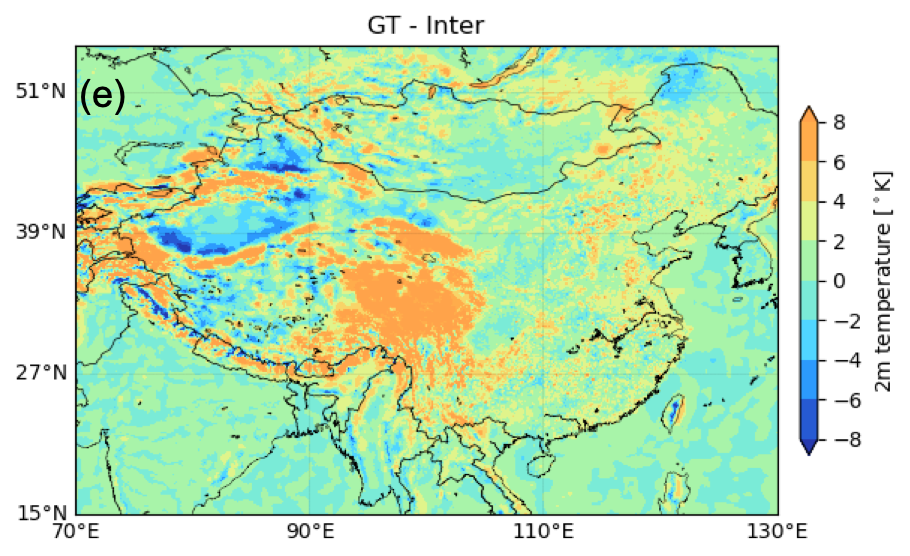}
    \end{subfigure}
         \hspace{-0.8cm}
    \begin{subfigure}[t]{0.27\textwidth}
        \includegraphics[width=\linewidth]{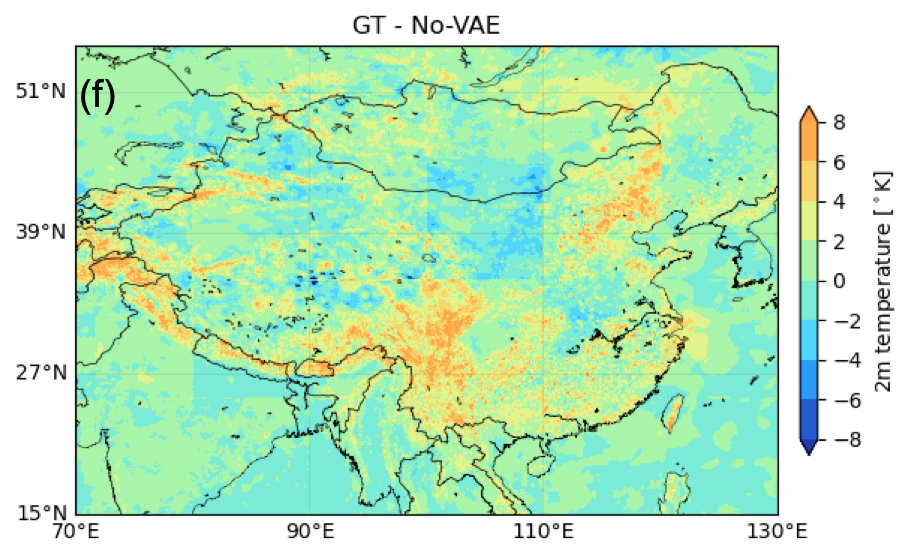}
    \end{subfigure}
    \hspace{-0.8cm}
    \begin{subfigure}[t]{0.27\textwidth} \includegraphics[width=\linewidth]{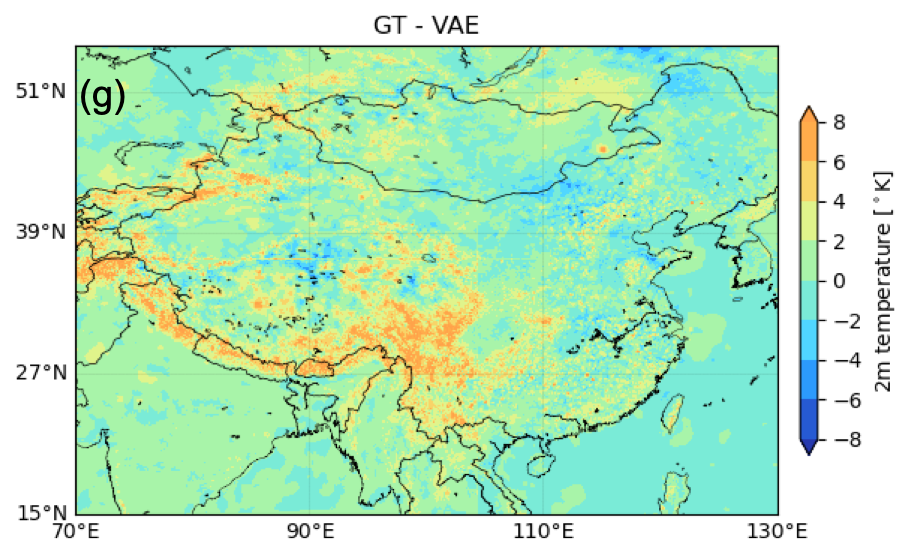}
    \end{subfigure}

    \begin{subfigure}[t]{0.27\textwidth}
        \includegraphics[width=\linewidth]{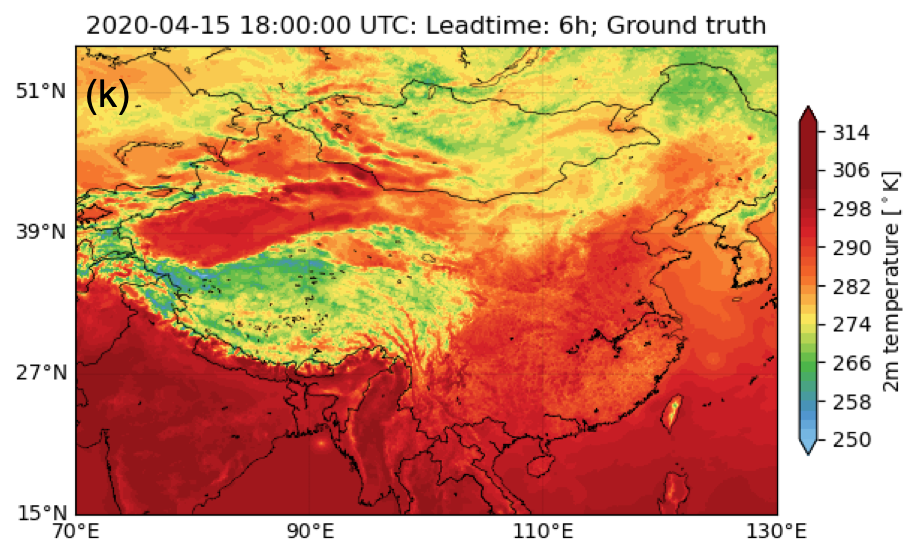}
    \end{subfigure}
    \hspace{-0.8cm}
    \begin{subfigure}[t]{0.27\textwidth}
        \includegraphics[width=\linewidth]{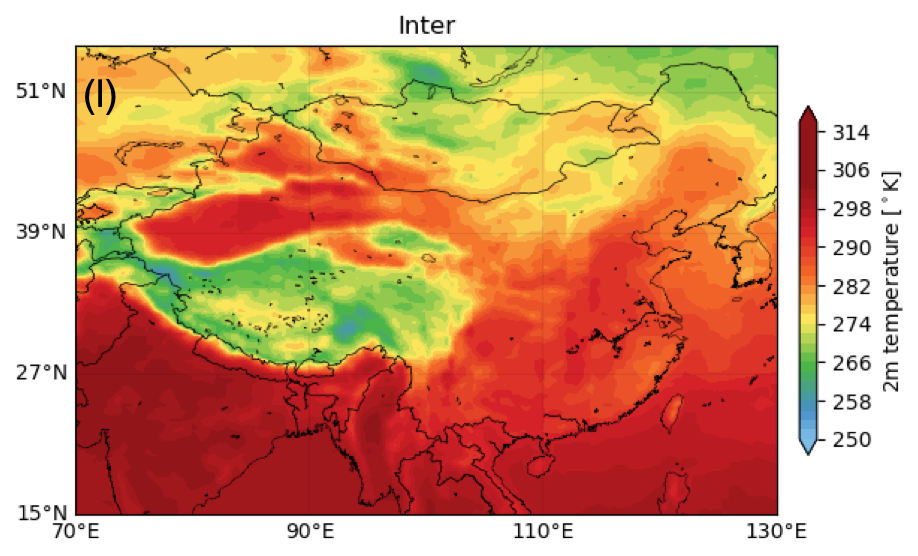}
    \end{subfigure}
     \hspace{-0.8cm}
    \begin{subfigure}[t]{0.27\textwidth}
        \includegraphics[width=\linewidth]{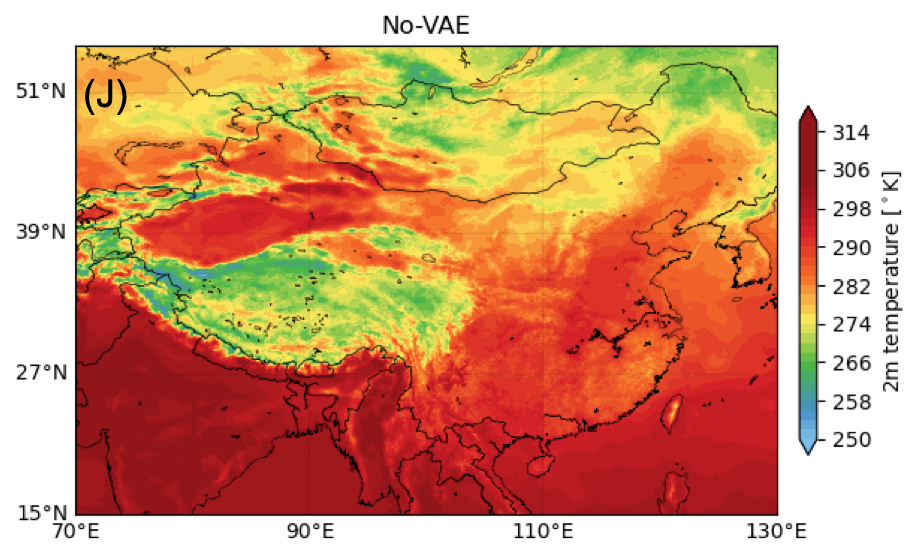}
    \end{subfigure}
    \hspace{-0.8cm}
    \begin{subfigure}[t]{0.27\textwidth}
        \includegraphics[width=\linewidth]{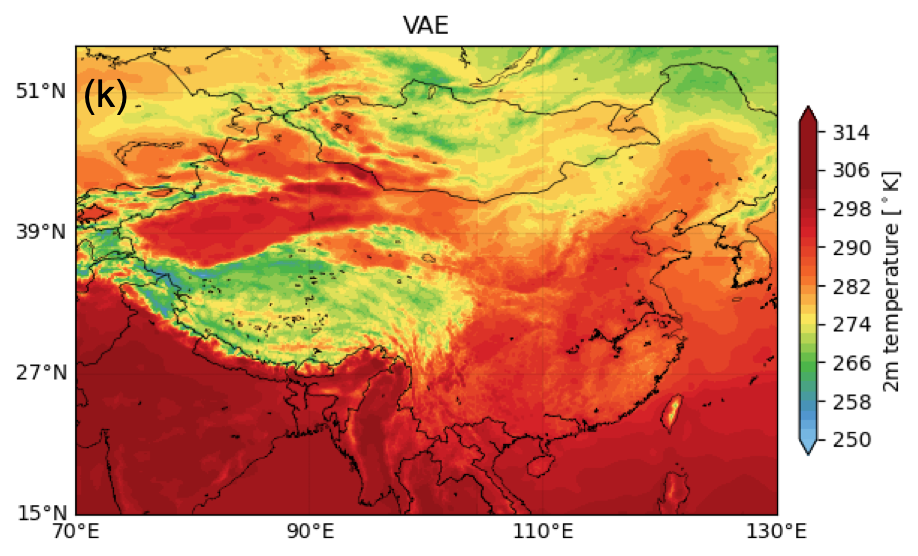}
    \end{subfigure}

    \begin{subfigure}[t]{0.27\textwidth}
        \includegraphics[width=\linewidth]{kongbai.png}
    \end{subfigure}
     \hspace{-0.8cm}
    \begin{subfigure}[t]{0.27\textwidth}
        \includegraphics[width=\linewidth]{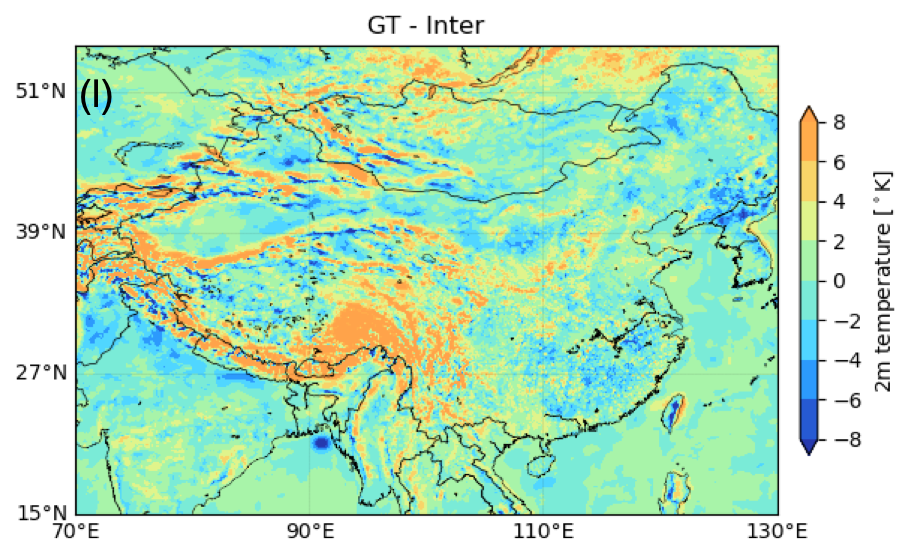}
    \end{subfigure}
         \hspace{-0.8cm}
    \begin{subfigure}[t]{0.27\textwidth}
        \includegraphics[width=\linewidth]{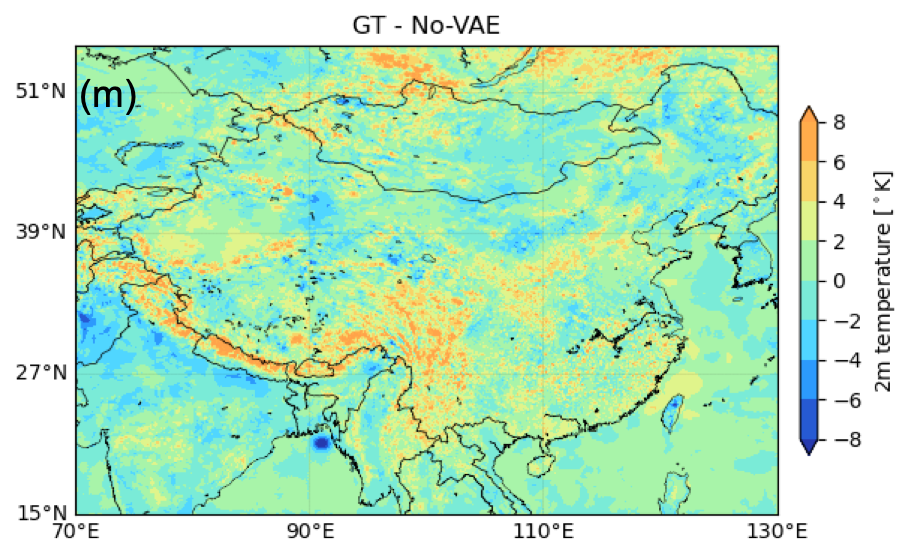}
    \end{subfigure}
    \hspace{-0.8cm}
    \begin{subfigure}[t]{0.27\textwidth} \includegraphics[width=\linewidth]{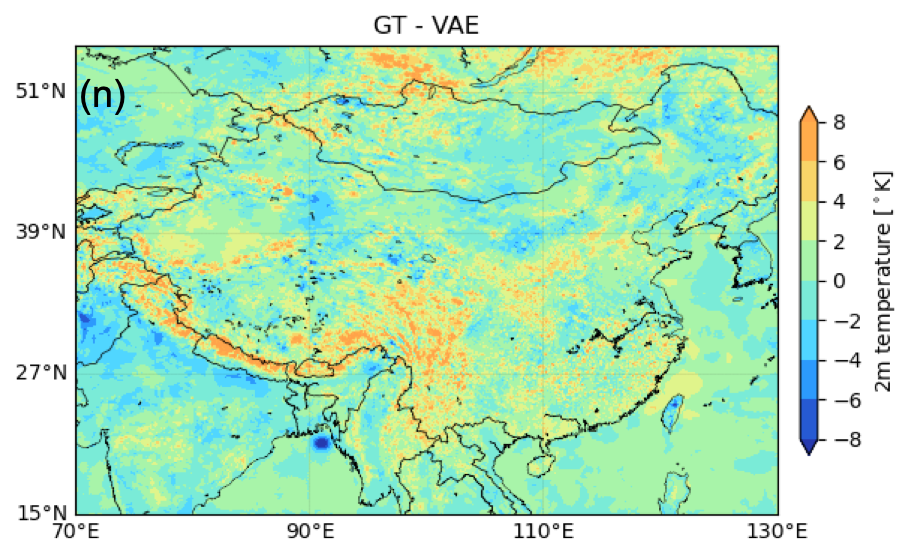}
    \end{subfigure}

    \begin{subfigure}[t]{0.27\textwidth}
        \includegraphics[width=\linewidth]{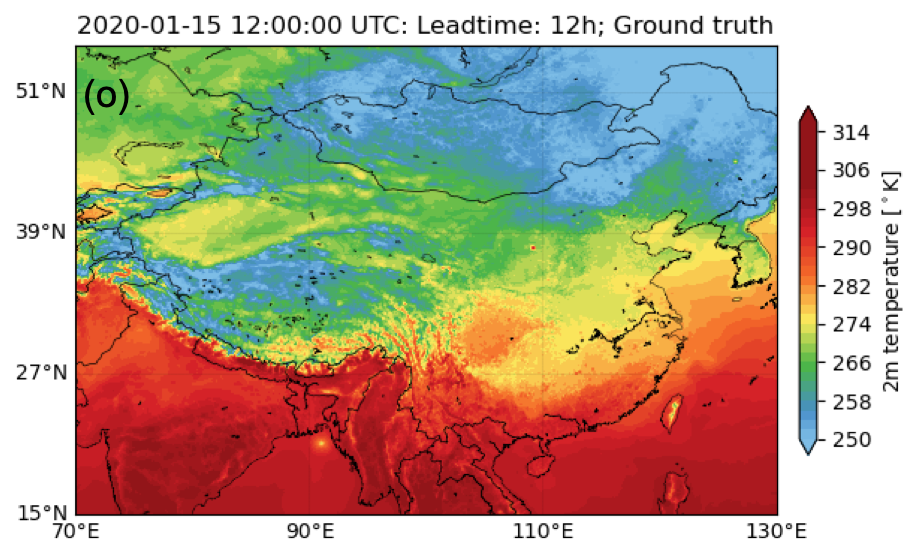}
    \end{subfigure}
    \hspace{-0.8cm}
    \begin{subfigure}[t]{0.27\textwidth}
        \includegraphics[width=\linewidth]{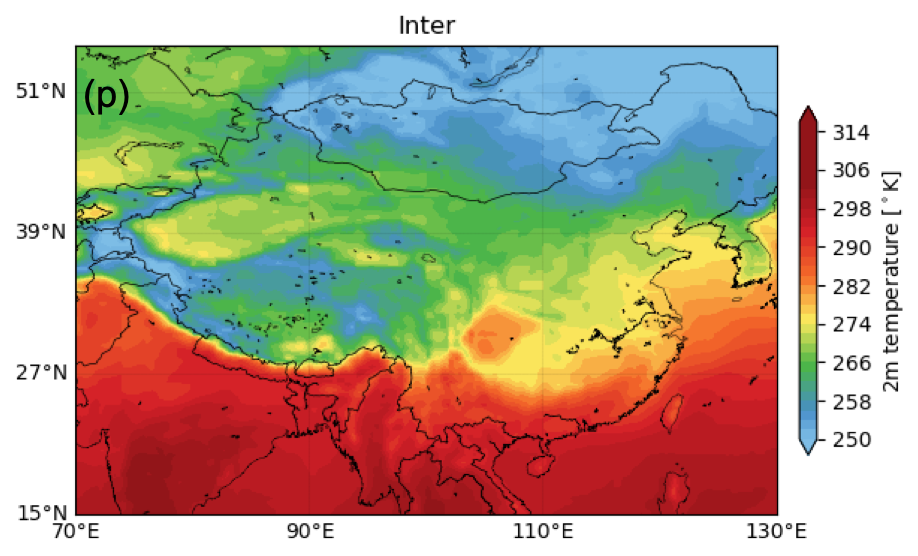}
    \end{subfigure}
     \hspace{-0.8cm}
    \begin{subfigure}[t]{0.27\textwidth}
        \includegraphics[width=\linewidth]{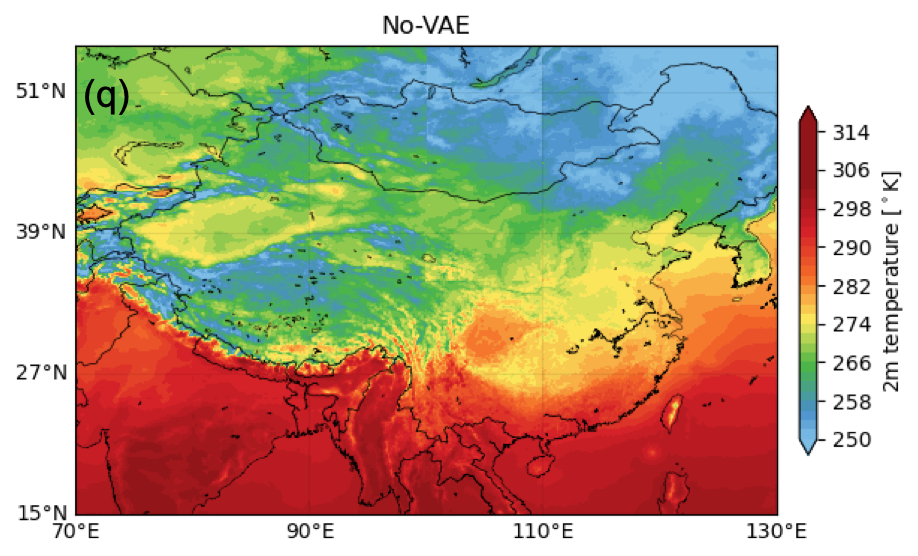}
    \end{subfigure}
    \hspace{-0.8cm}
    \begin{subfigure}[t]{0.27\textwidth}
        \includegraphics[width=\linewidth]{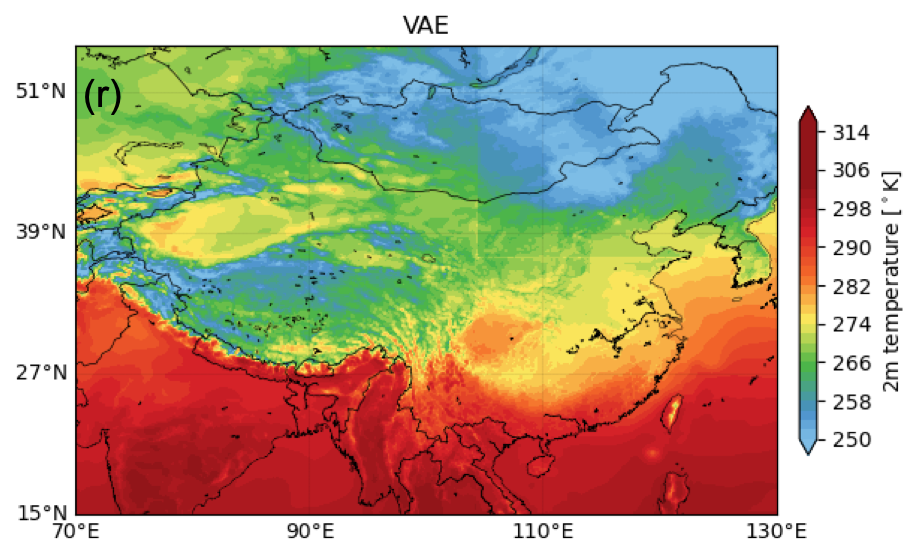}
    \end{subfigure}

    \begin{subfigure}[t]{0.27\textwidth}
        \includegraphics[width=\linewidth]{kongbai.png}
    \end{subfigure}
     \hspace{-0.8cm}
    \begin{subfigure}[t]{0.27\textwidth}
        \includegraphics[width=\linewidth]{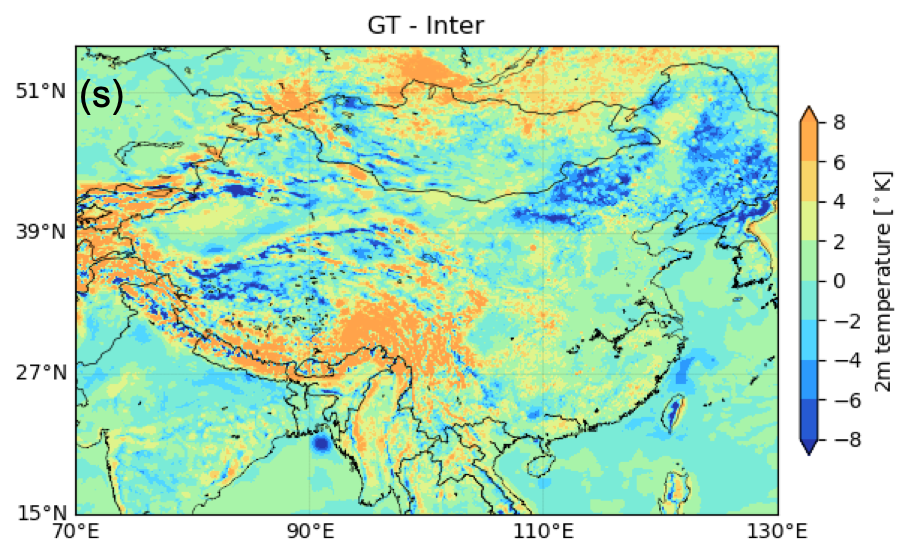}
    \end{subfigure}
         \hspace{-0.8cm}
    \begin{subfigure}[t]{0.27\textwidth}
        \includegraphics[width=\linewidth]{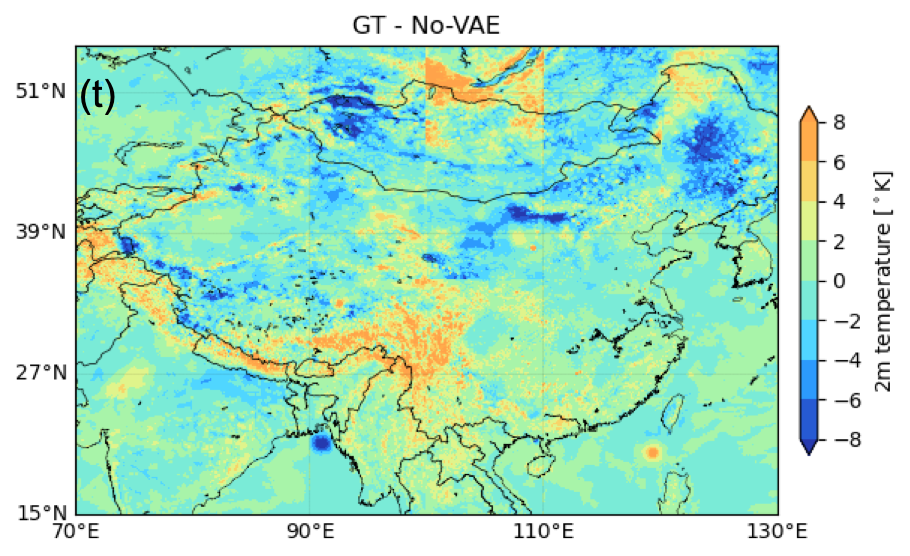}
    \end{subfigure}
    \hspace{-0.8cm}
    \begin{subfigure}[t]{0.27\textwidth} \includegraphics[width=\linewidth]{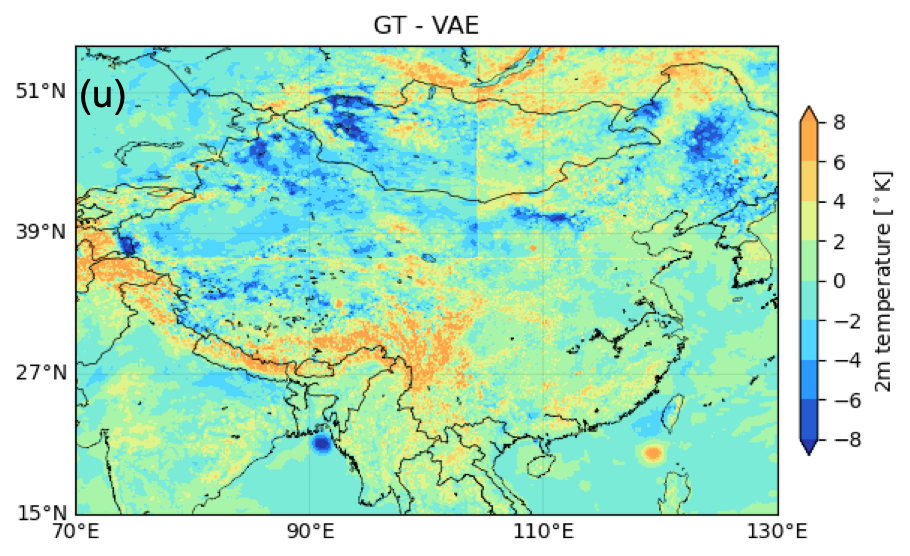}
    \end{subfigure}
    \caption{Example downscaling for 2~m temperature with a lead time of 1~h (panels: b - d), 6~h (panels: i-k) and 12~h (p-f), by interpretation, No-VAE, and VAE methods respectively. (a,h,o) are the ground truth from the HRCLAS dataset. (e-g, l-n, s-u) The difference between the ground truth and the downscaled fields generated by interpretation, No-VAE, and VAE method at the time of 1\ts{st} November 2019, 01:00 UTC,  15\ts{st} April 2020, 18:00 UTC, and 15\ts{st} December 2020, 12:00 UTC. 
}
    \label{fig:sample1}
\end{figure*}

Furthermore, Fig.~\ref{fig:ps} presents the average zonal power spectrum plots for the No-VAE and VAE downscaling models at a 1-hour lead time, compared to the ground truth HRCLDAS data and the baseline interpolation method. The findings reveal that the baseline interpolation method fails to capture significant details across all scales for the three variables. This is expected, as interpolation tends to average out variances, leading to a smoother appearance, as visually confirmed in Fig ~\ref{fig:sample1} and Fig.~ \ref{fig:sample2}. In contrast, both the VAE and No-VAE models preserve more details, with their performances being quite similar. This suggests that the compact data representation from the VAE does not lead to substantial information loss in our proof-of-concept downscaling task compared to using the original HRCLDAS data.

It is worth mentioning that the No-VAE model generates artifacts at small scales for T$_{2M}$ as the examples demonstrated in Fig.~\ref{fig:sample1}c and Fig.~\ref{fig:sample1}f. Sudden temperature changes create strong contrasts in specific areas, such as northeast China. This issue arises from the unpatching process, where small patches are reconstructed into the original image due to computational limitations. In contrast, this problem can be effectively mitigated by feeding the U-Net the entire latent features of HRCLDAS data directly from the VAE model, resulting in a smoother downscaling field, as demonstrated in Fig.~\ref{fig:sample1}d and Fig.~\ref{fig:sample1}g. 

In addition, we observed that both the VAE and No-VAE models struggle to capture fine-scale features of T$_{2M}$. This limitation is likely due to the significant elevation variations in the Himalayas and certain regions of China, where topography is crucial for maintaining local temperature gradients. Since the primary focus of this study is to validate the VAE compression method, we did not incorporate topography data in the current models. However, this will be considered in future research.

\begin{figure}[htb!]
    \centering
    \includegraphics[width=0.6\textwidth,height=0.4\textwidth]{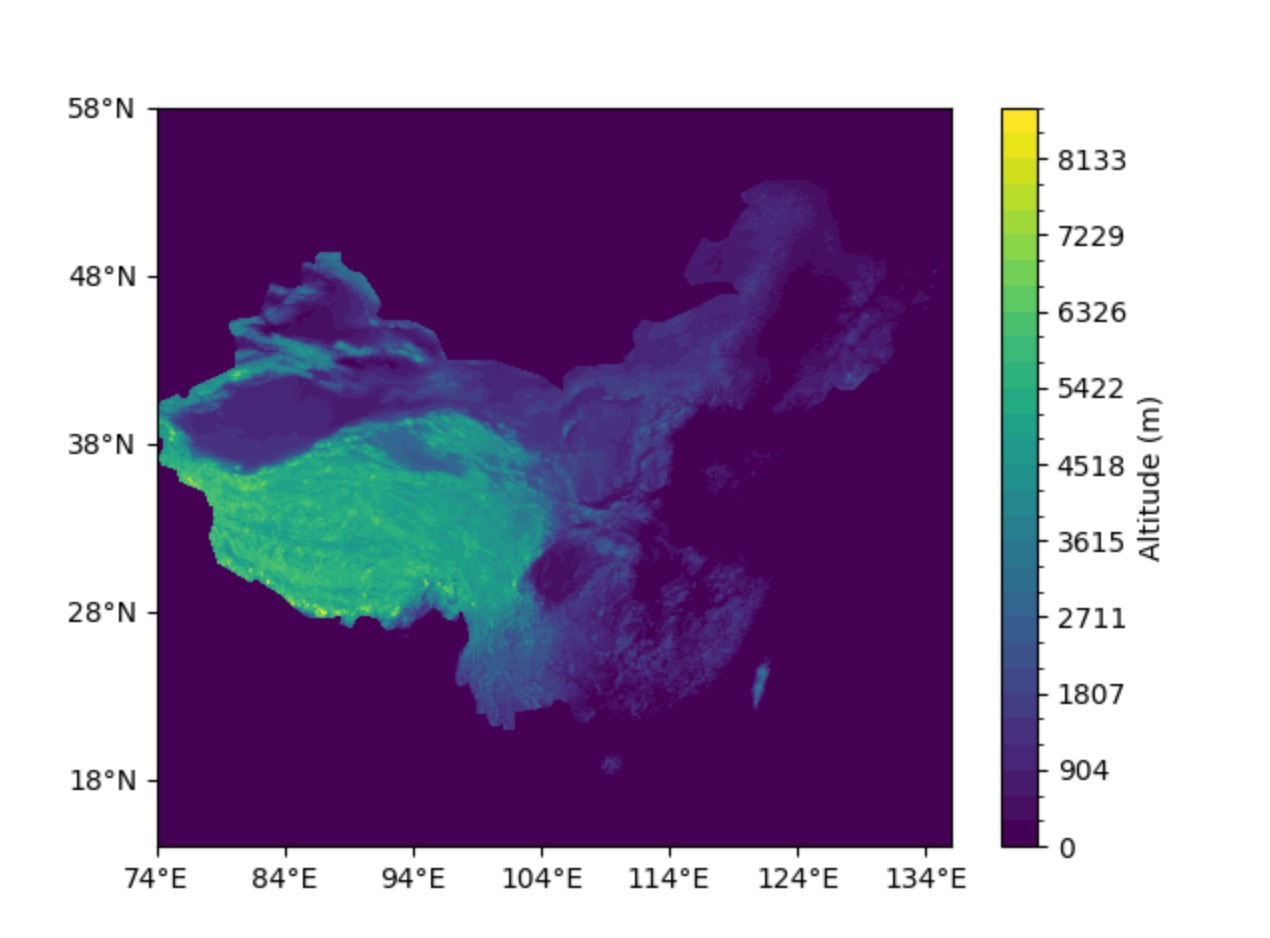}
    \caption{Topography height of the surface in the target region.}
    \label{fig:top}
\end{figure}

\begin{figure*}[ht!]

    \begin{subfigure}[t]{0.27\textwidth}
        \includegraphics[width=\linewidth]{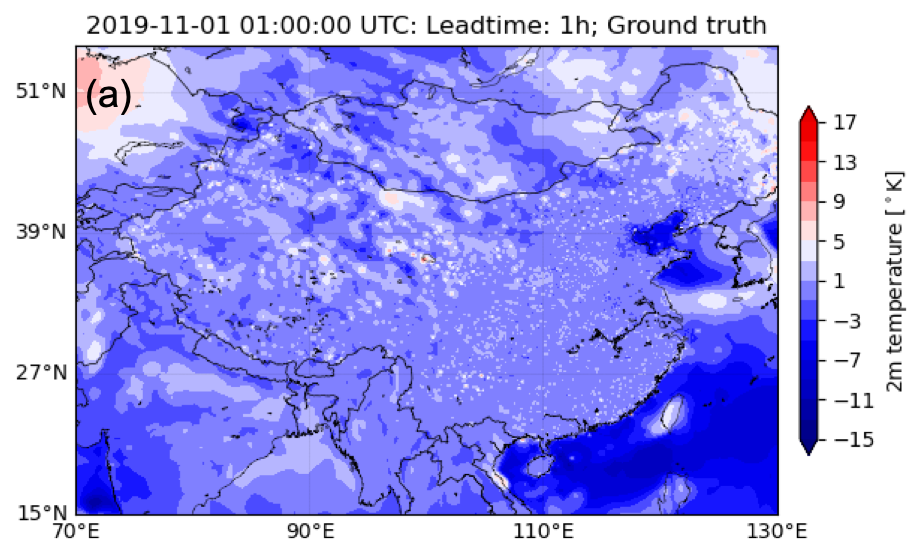}
    \end{subfigure}
    \hspace{-0.82cm}
    \begin{subfigure}[t]{0.27\textwidth}
        \includegraphics[width=\linewidth]{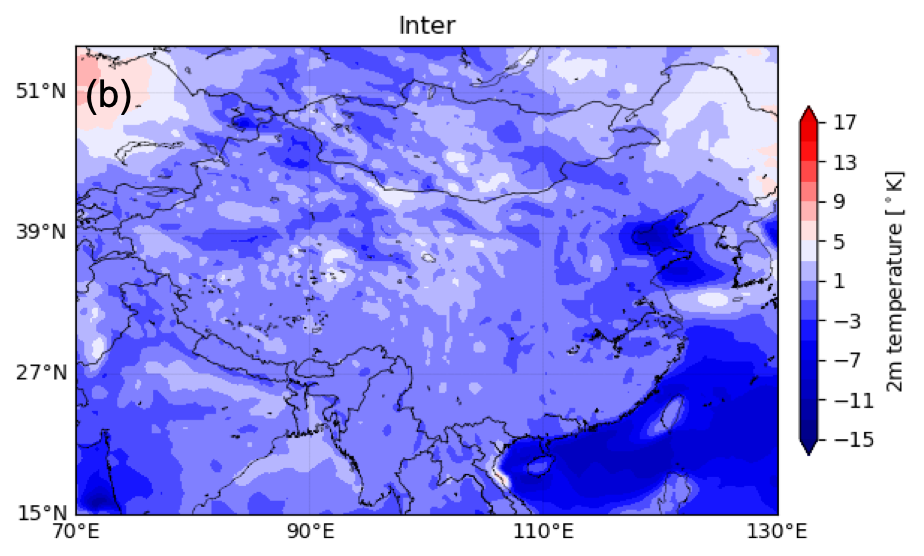}
    \end{subfigure}
     \hspace{-0.82cm}
    \begin{subfigure}[t]{0.27\textwidth}
        \includegraphics[width=\linewidth]{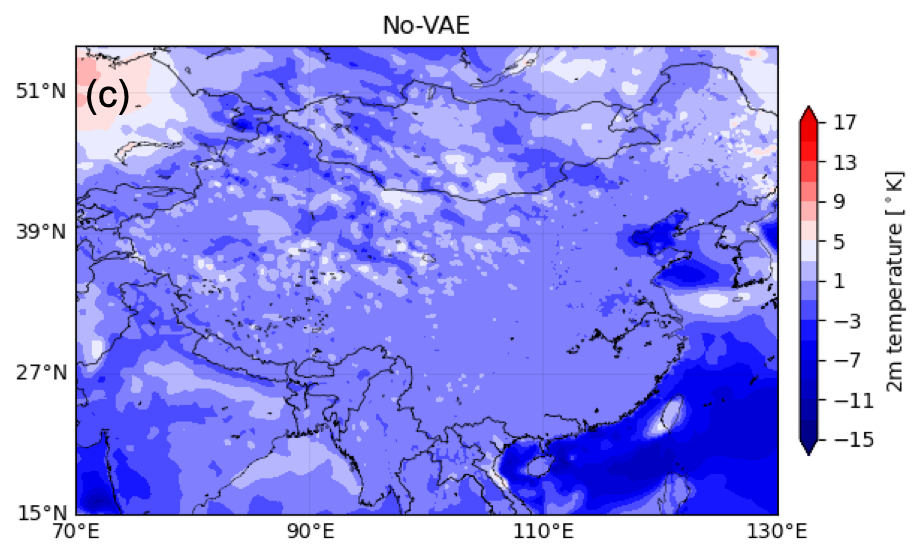}
    \end{subfigure}
    \hspace{-0.82cm}
    \begin{subfigure}[t]{0.27\textwidth}
        \includegraphics[width=\linewidth]{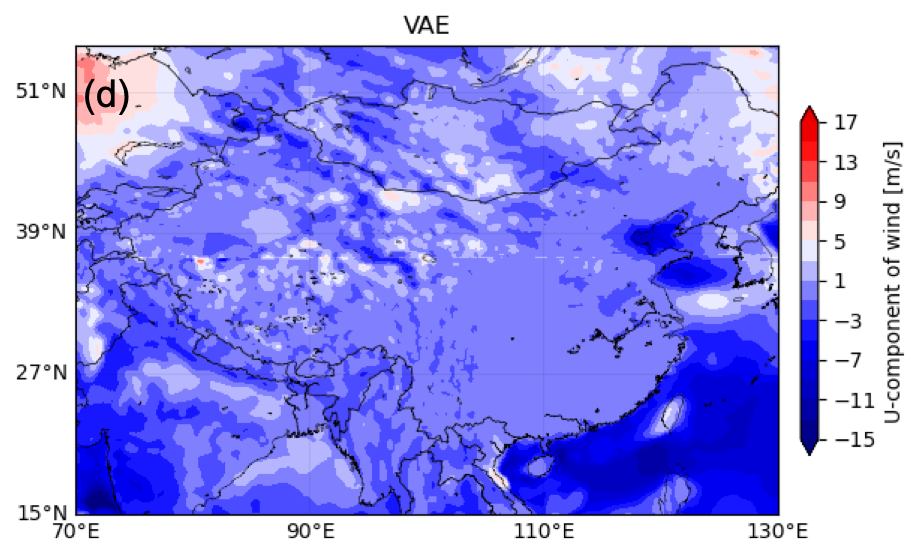}
    \end{subfigure}

    \begin{subfigure}[t]{0.27\textwidth}
        \includegraphics[width=\linewidth]{kongbai.png}
    \end{subfigure}
     \hspace{-0.8cm}
    \begin{subfigure}[t]{0.27\textwidth}
        \includegraphics[width=\linewidth]{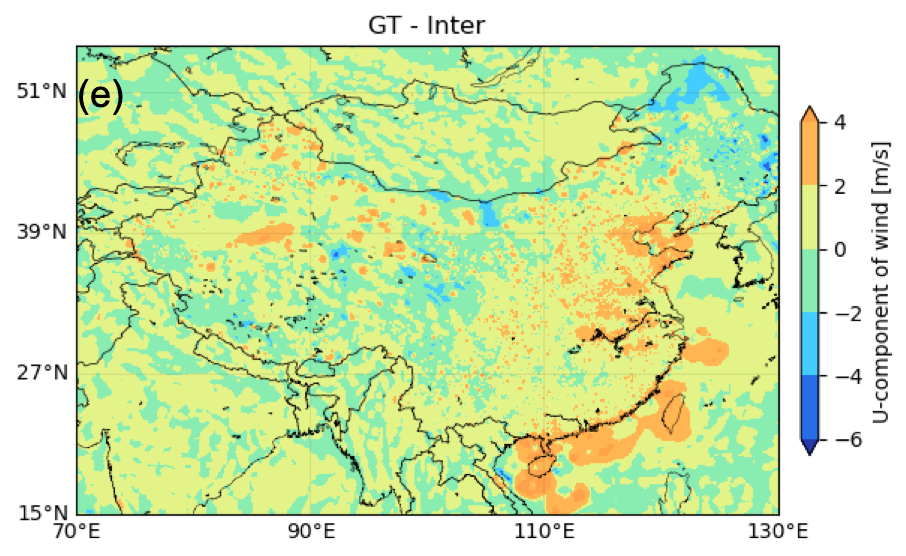}
    \end{subfigure}
         \hspace{-0.8cm}
    \begin{subfigure}[t]{0.27\textwidth}
        \includegraphics[width=\linewidth]{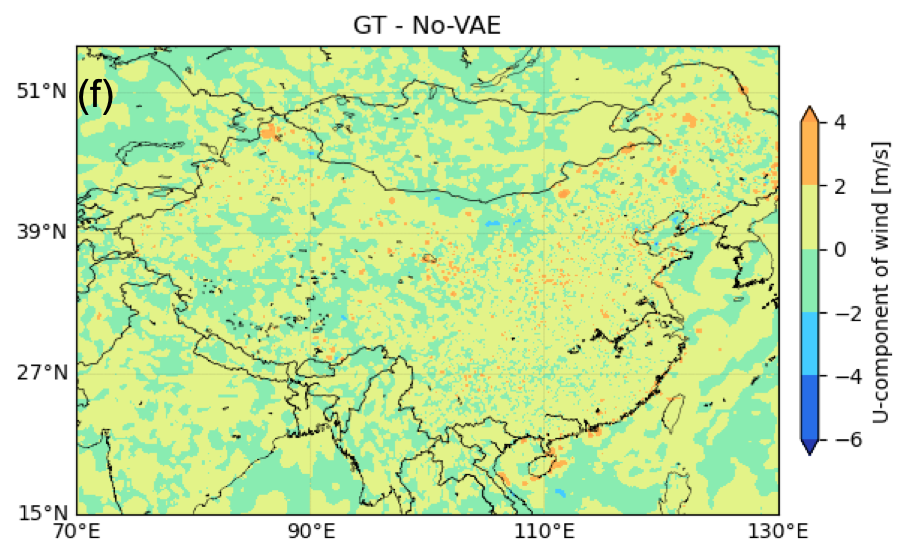}
    \end{subfigure}
    \hspace{-0.8cm}
    \begin{subfigure}[t]{0.27\textwidth} \includegraphics[width=\linewidth]{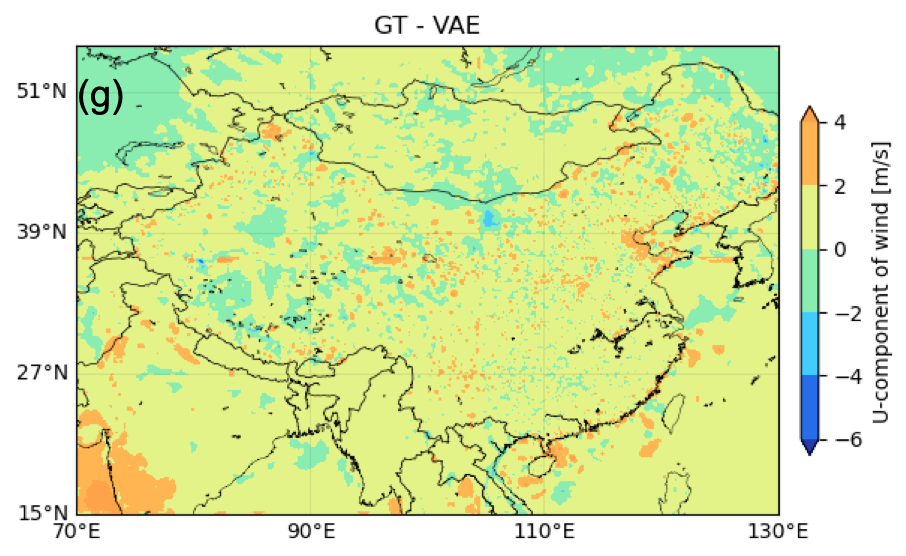}
    \end{subfigure}

    \begin{subfigure}[t]{0.27\textwidth}
        \includegraphics[width=\linewidth]{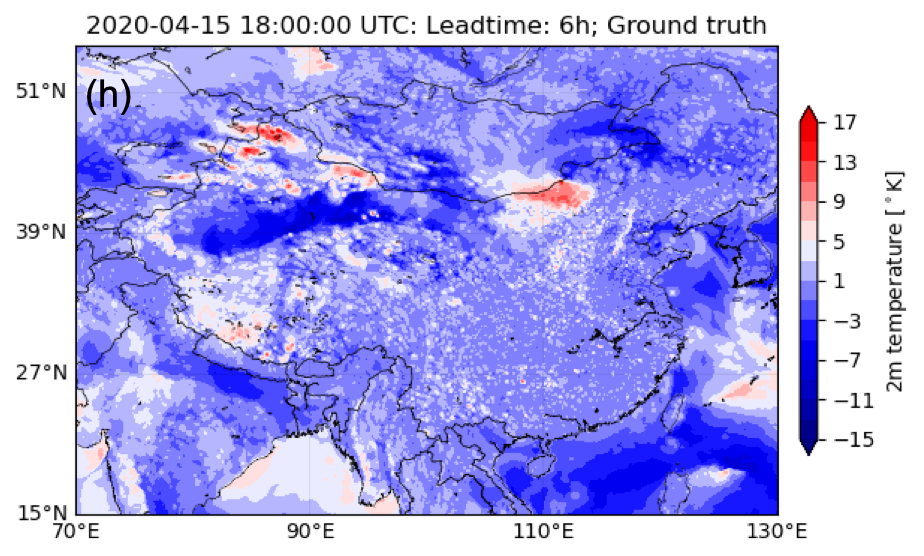}
    \end{subfigure}
    \hspace{-0.82cm}
    \begin{subfigure}[t]{0.27\textwidth}
        \includegraphics[width=\linewidth]{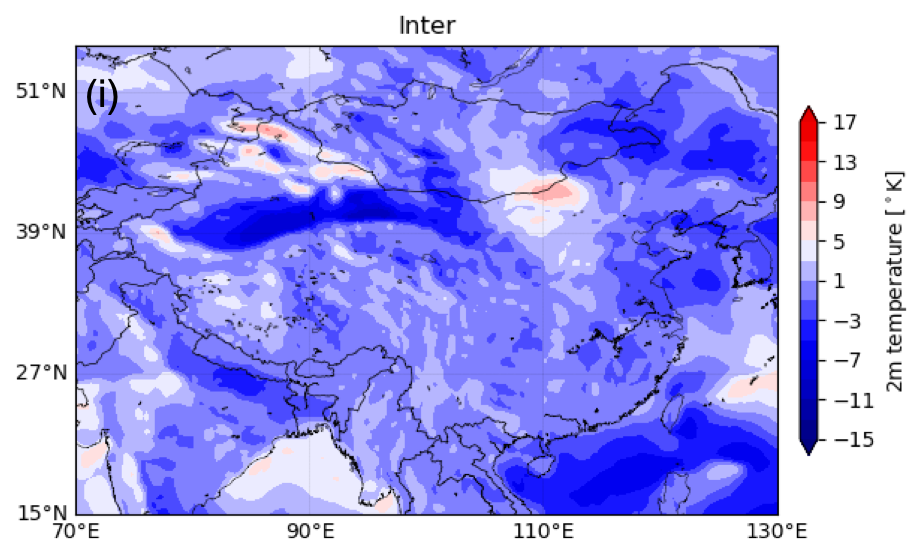}
    \end{subfigure}
     \hspace{-0.82cm}
    \begin{subfigure}[t]{0.27\textwidth}
        \includegraphics[width=\linewidth]{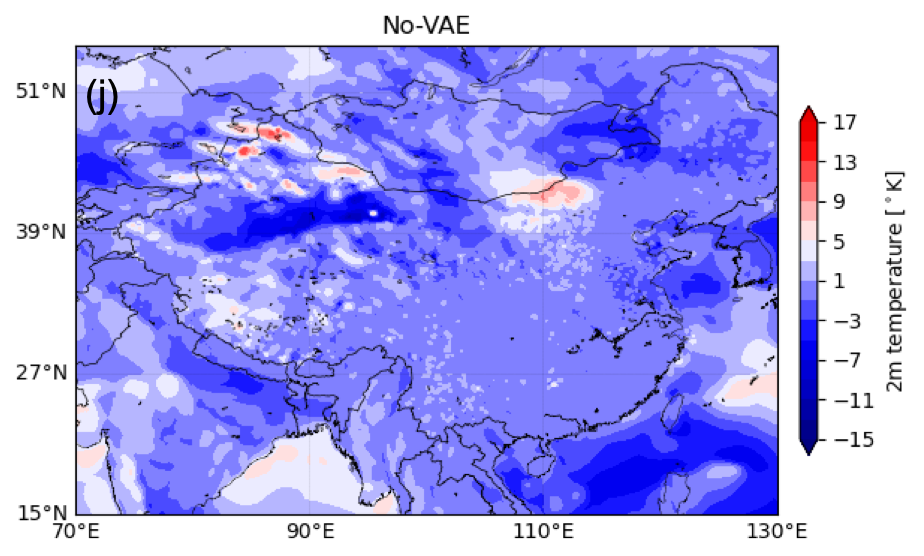}
    \end{subfigure}
    \hspace{-0.82cm}
    \begin{subfigure}[t]{0.27\textwidth}
        \includegraphics[width=\linewidth]{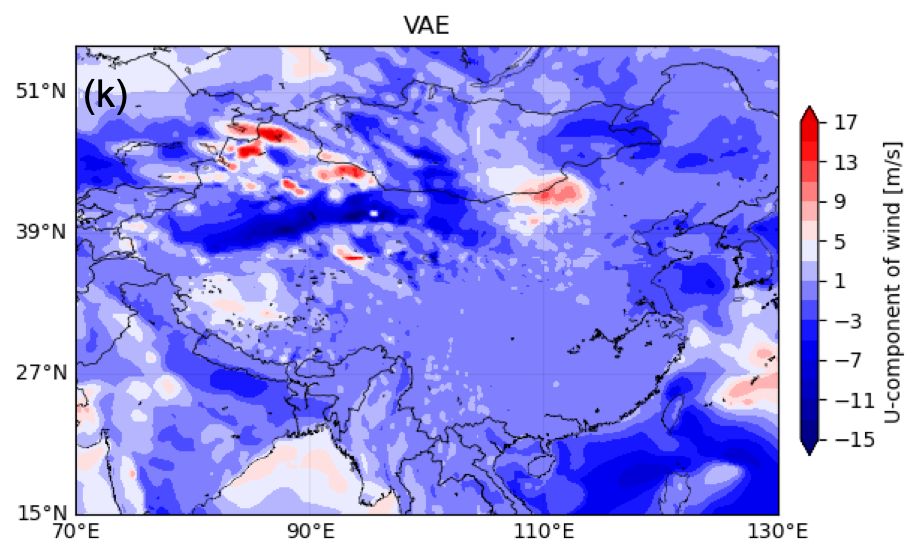}
    \end{subfigure}

    \begin{subfigure}[t]{0.27\textwidth}
        \includegraphics[width=\linewidth]{kongbai.png}
    \end{subfigure}
     \hspace{-0.82cm}
    \begin{subfigure}[t]{0.27\textwidth}
        \includegraphics[width=\linewidth]{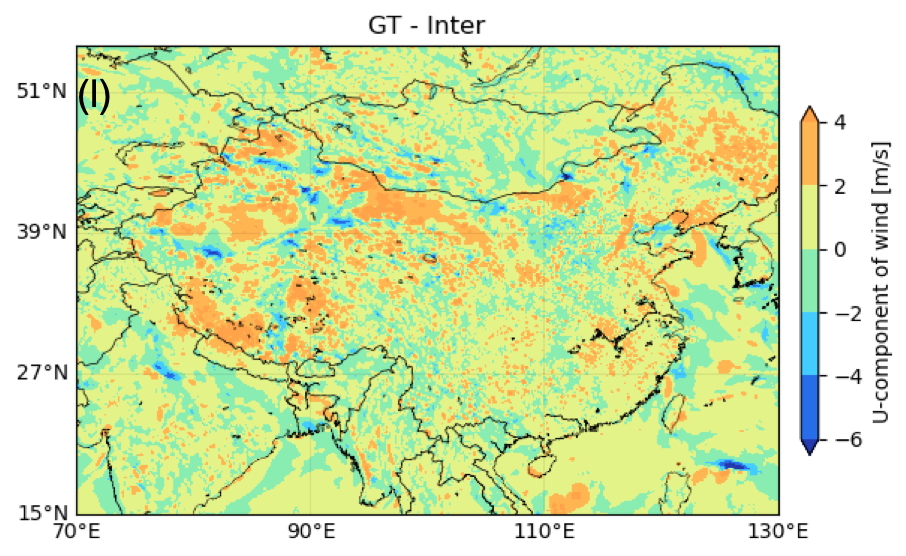}
    \end{subfigure}
         \hspace{-0.82cm}
    \begin{subfigure}[t]{0.27\textwidth}
        \includegraphics[width=\linewidth]{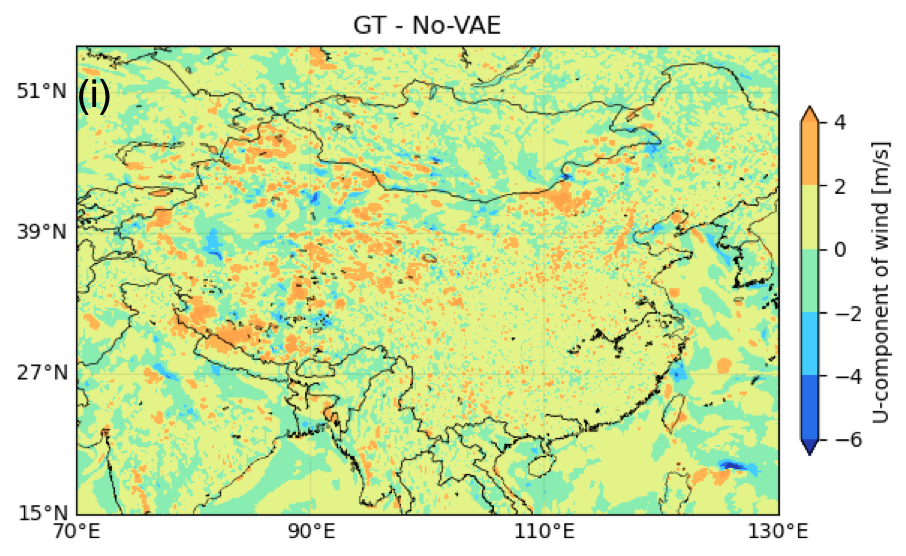}
    \end{subfigure}
    \hspace{-0.82cm}
    \begin{subfigure}[t]{0.27\textwidth} \includegraphics[width=\linewidth]{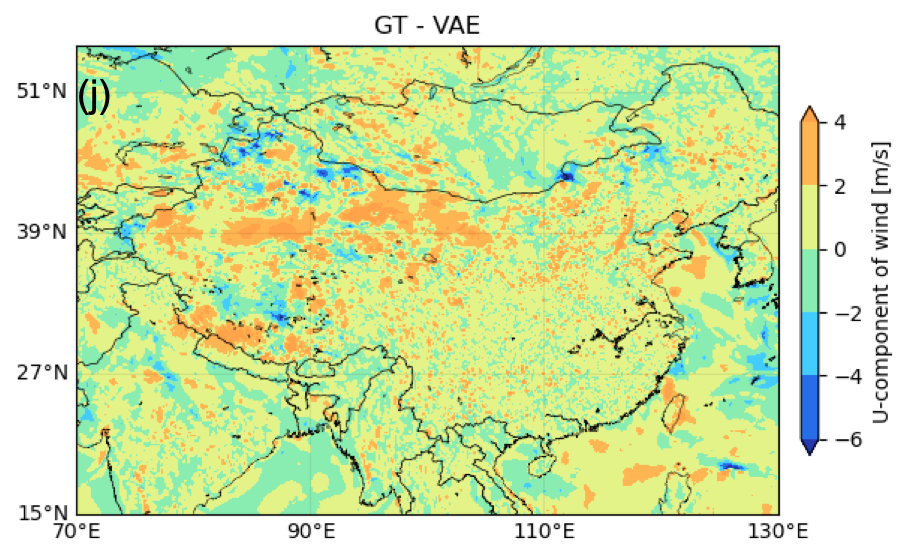}
    \end{subfigure}

    \begin{subfigure}[t]{0.27\textwidth}
        \includegraphics[width=\linewidth]{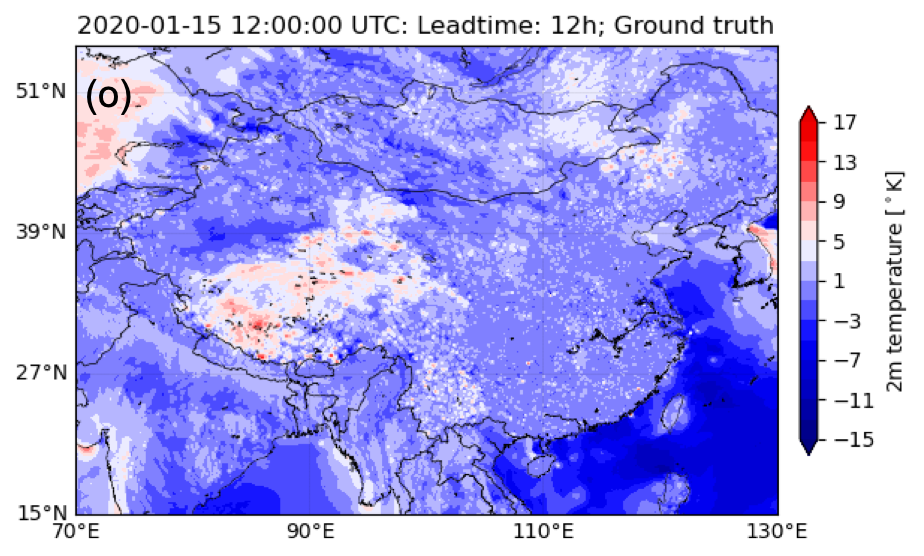}
    \end{subfigure}
    \hspace{-0.82cm}
    \begin{subfigure}[t]{0.27\textwidth}
        \includegraphics[width=\linewidth]{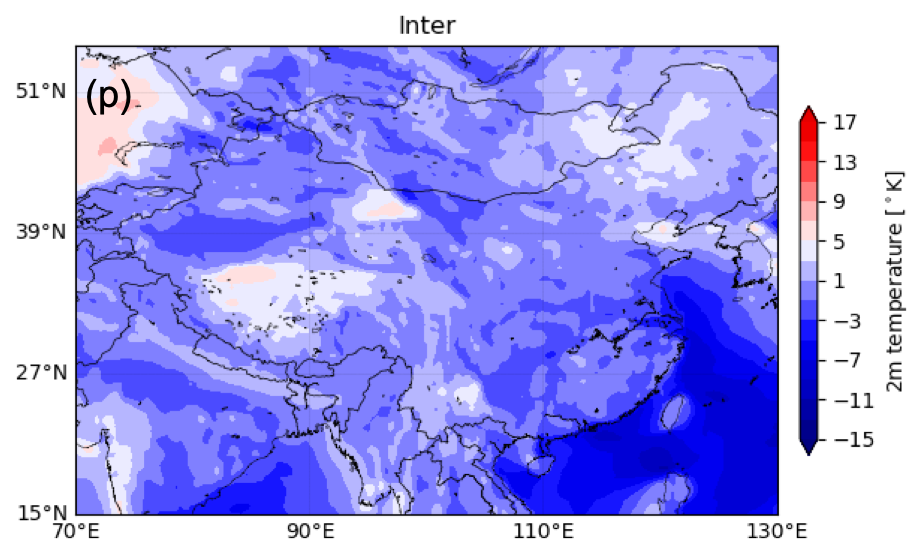}
    \end{subfigure}
     \hspace{-0.82cm}
    \begin{subfigure}[t]{0.27\textwidth}
        \includegraphics[width=\linewidth]{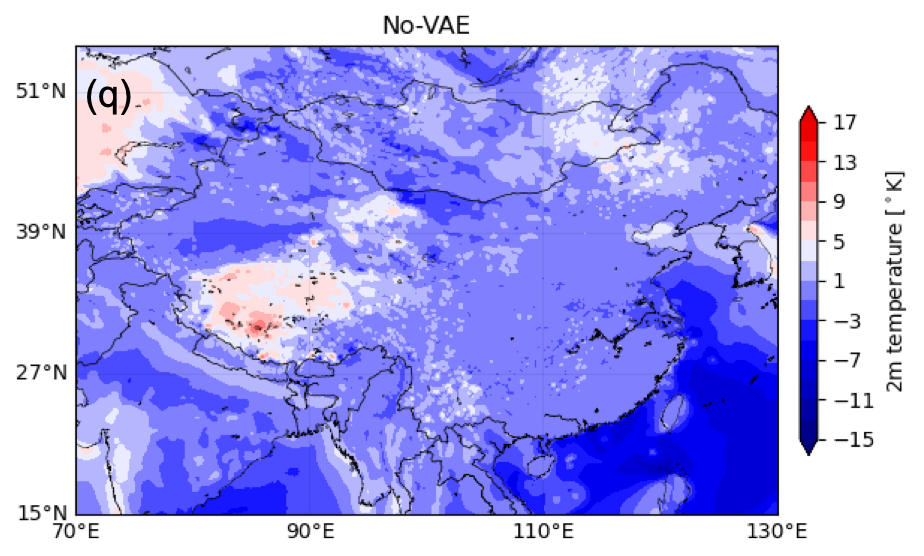}
    \end{subfigure}
    \hspace{-0.82cm}
    \begin{subfigure}[t]{0.27\textwidth}
        \includegraphics[width=\linewidth]{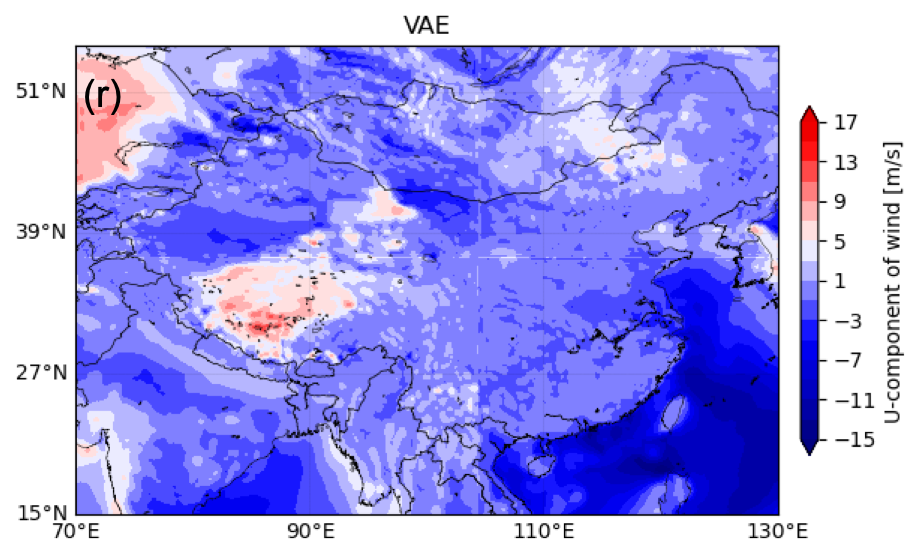}
    \end{subfigure}

    \begin{subfigure}[t]{0.27\textwidth}
        \includegraphics[width=\linewidth]{kongbai.png}
    \end{subfigure}
     \hspace{-0.82cm}
    \begin{subfigure}[t]{0.27\textwidth}
        \includegraphics[width=\linewidth]{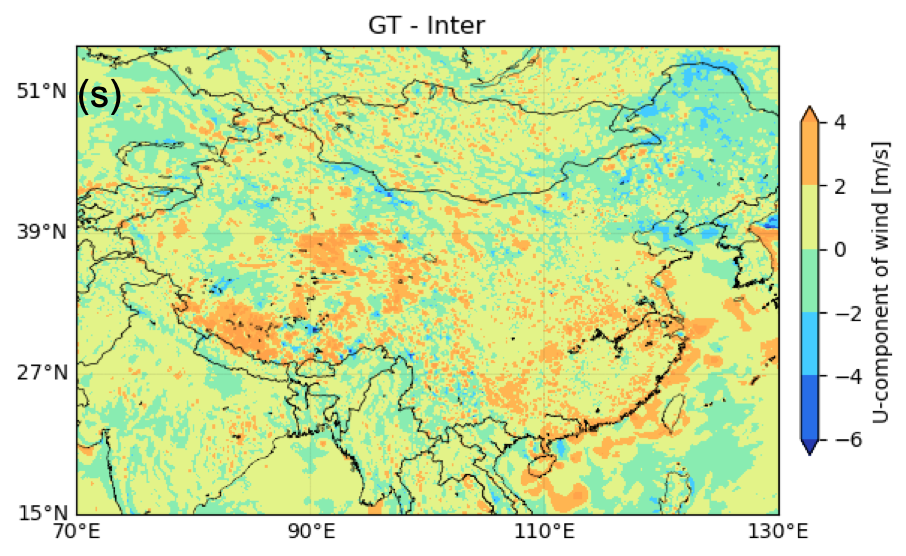}
    \end{subfigure}
         \hspace{-0.82cm}
    \begin{subfigure}[t]{0.27\textwidth}
        \includegraphics[width=\linewidth]{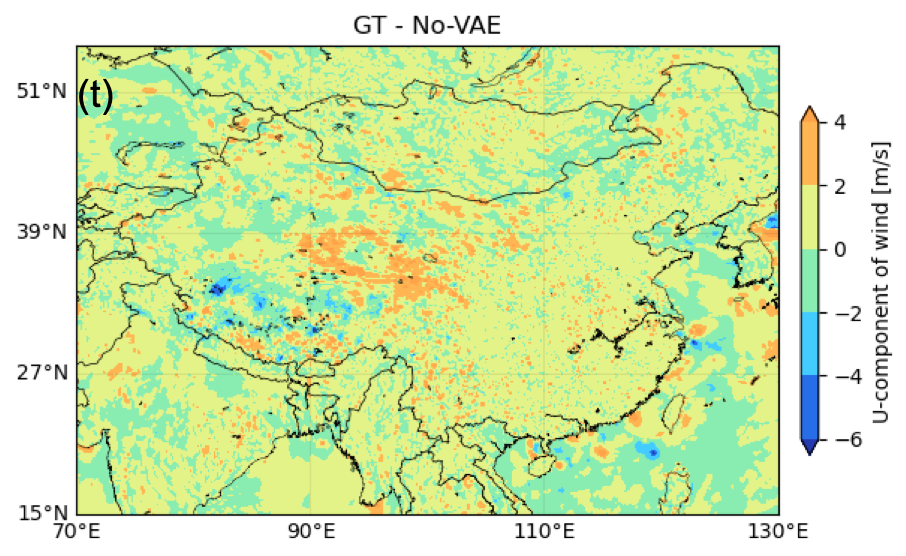}
    \end{subfigure}
    \hspace{-0.82cm}
    \begin{subfigure}[t]{0.27\textwidth} \includegraphics[width=\linewidth]{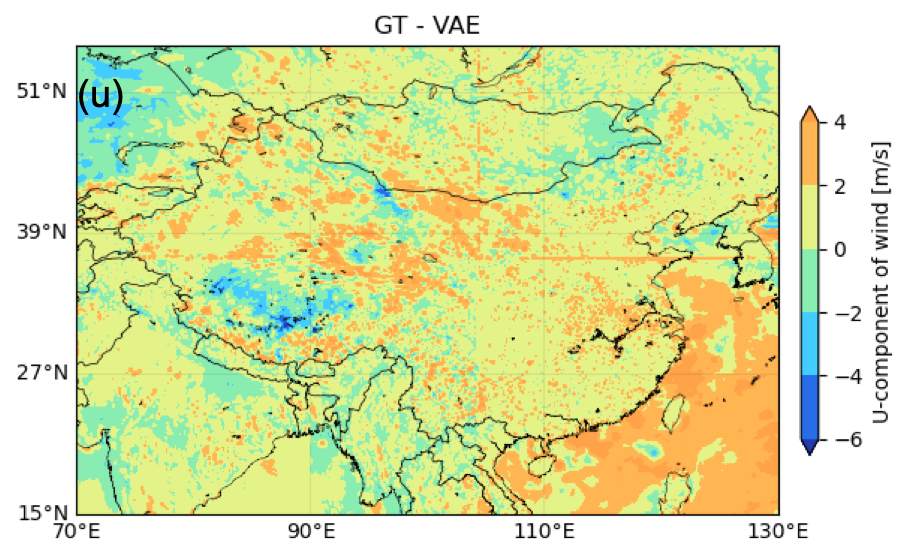}
    \end{subfigure}

    \caption{Example downscaling for U-component of wind with a lead time of 1~h (panels: b - d), 6~h (panels: i-k) and 12~h (p-f), by interpretation, No-VAE, and VAE methods respectively. (a,h,o) are the ground truth from the HRCLAS dataset. (e-g, l-n, s-u) The difference between the ground truth and the downscaled fields generated by interpretation, No-VAE, and VAE method at the time of 1\ts{st} November 2019, 01:00 UTC,  15\ts{st} April 2020, 18:00 UTC, and 15\ts{st} December 2020, 12:00 UTC.}
    \label{fig:sample2}
\end{figure*}

\begin{figure*}[ht!]

    \begin{subfigure}[t]{0.27\textwidth}
        \includegraphics[width=\linewidth]{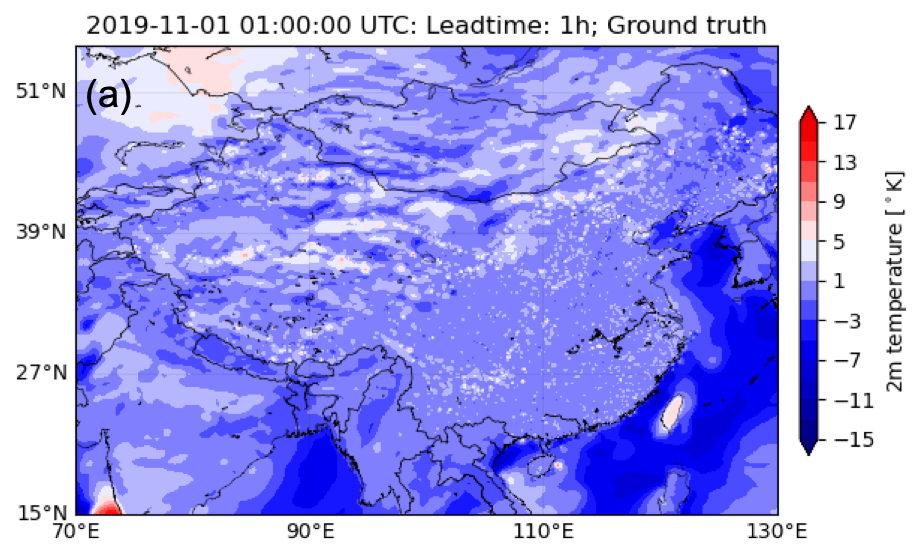}
    \end{subfigure}
    \hspace{-0.82cm}
    \begin{subfigure}[t]{0.27\textwidth}
        \includegraphics[width=\linewidth]{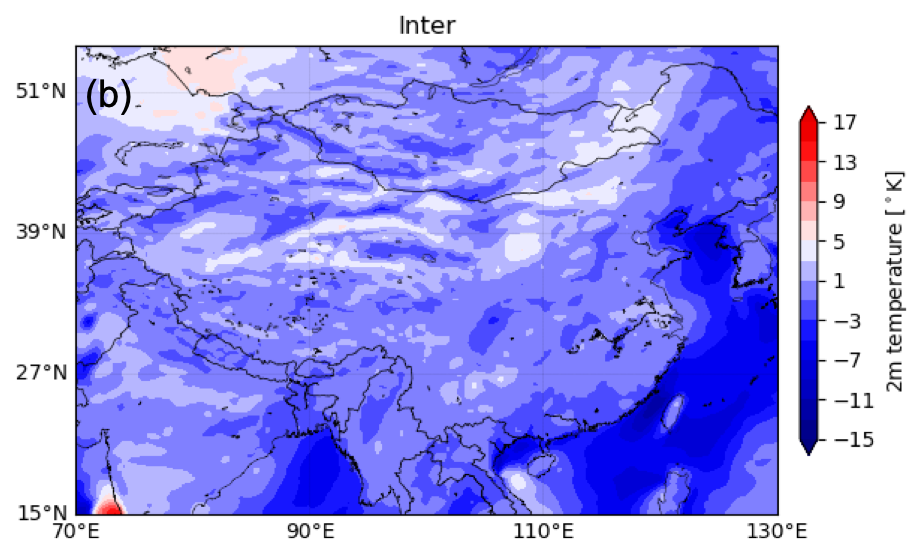}
    \end{subfigure}
     \hspace{-0.82cm}
    \begin{subfigure}[t]{0.27\textwidth}
        \includegraphics[width=\linewidth]{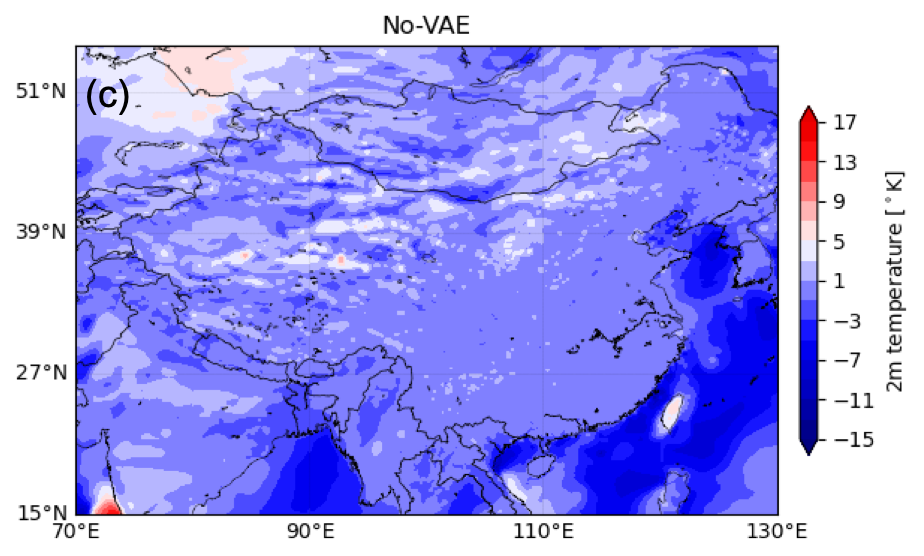}
    \end{subfigure}
    \hspace{-0.82cm}
    \begin{subfigure}[t]{0.27\textwidth}
        \includegraphics[width=\linewidth]{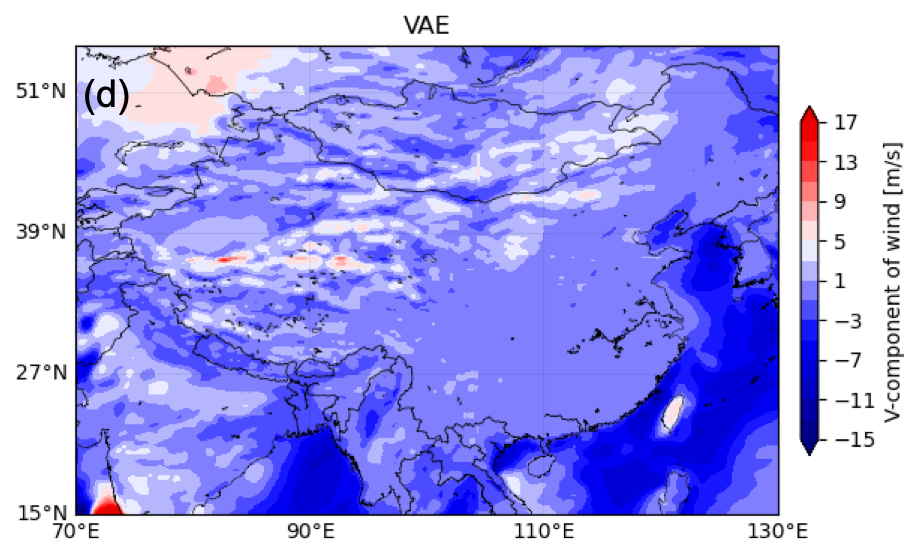}
    \end{subfigure}

    \begin{subfigure}[t]{0.27\textwidth}
        \includegraphics[width=\linewidth]{kongbai.png}
    \end{subfigure}
     \hspace{-0.82cm}
    \begin{subfigure}[t]{0.27\textwidth}
        \includegraphics[width=\linewidth]{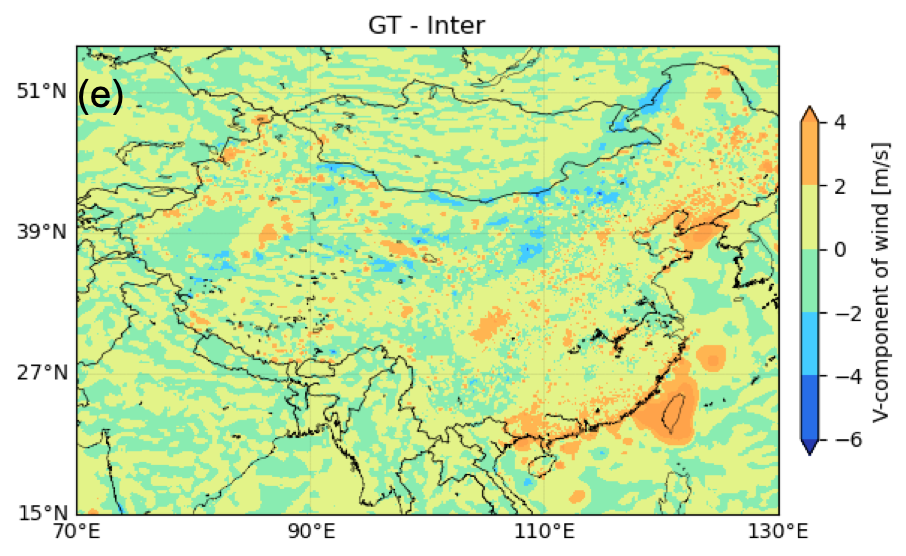}
    \end{subfigure}
         \hspace{-0.82cm}
    \begin{subfigure}[t]{0.27\textwidth}
        \includegraphics[width=\linewidth]{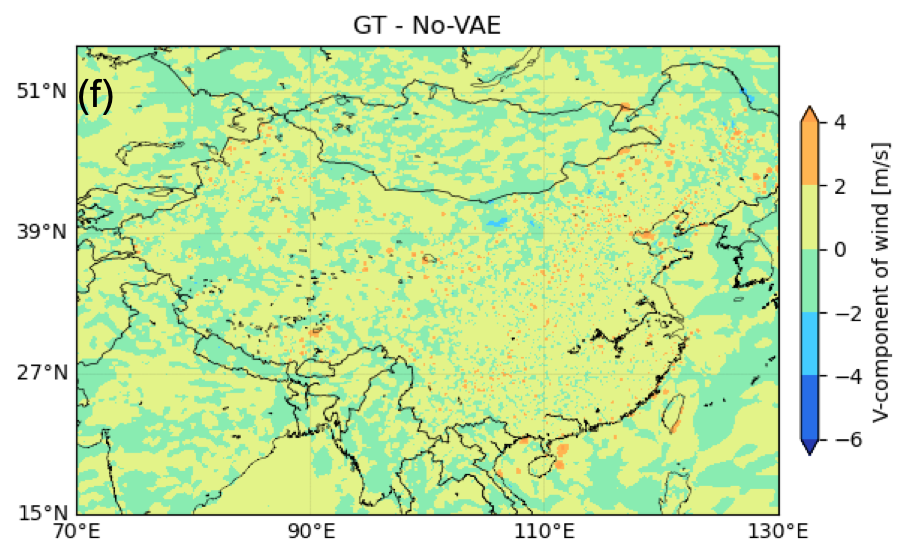}
    \end{subfigure}
    \hspace{-0.82cm}
    \begin{subfigure}[t]{0.27\textwidth} \includegraphics[width=\linewidth]{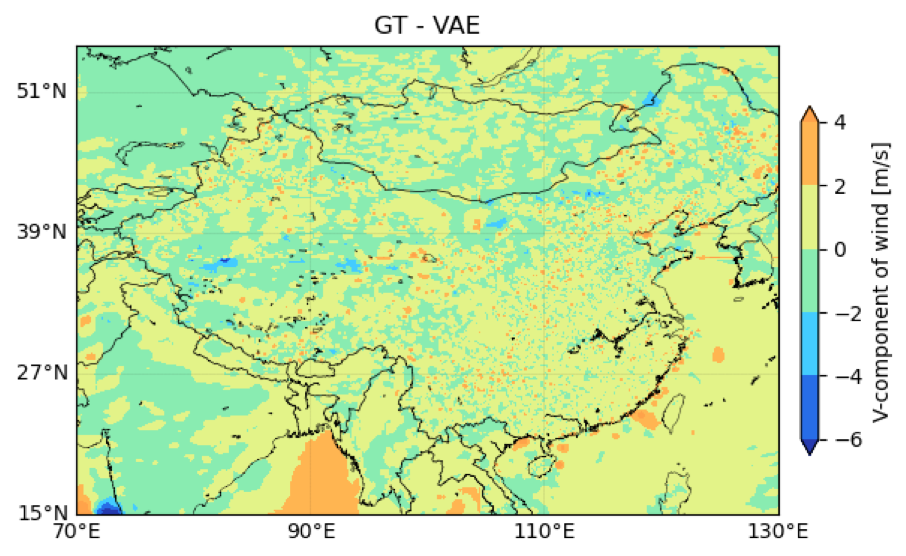}
    \end{subfigure}

    \begin{subfigure}[t]{0.27\textwidth}
        \includegraphics[width=\linewidth]{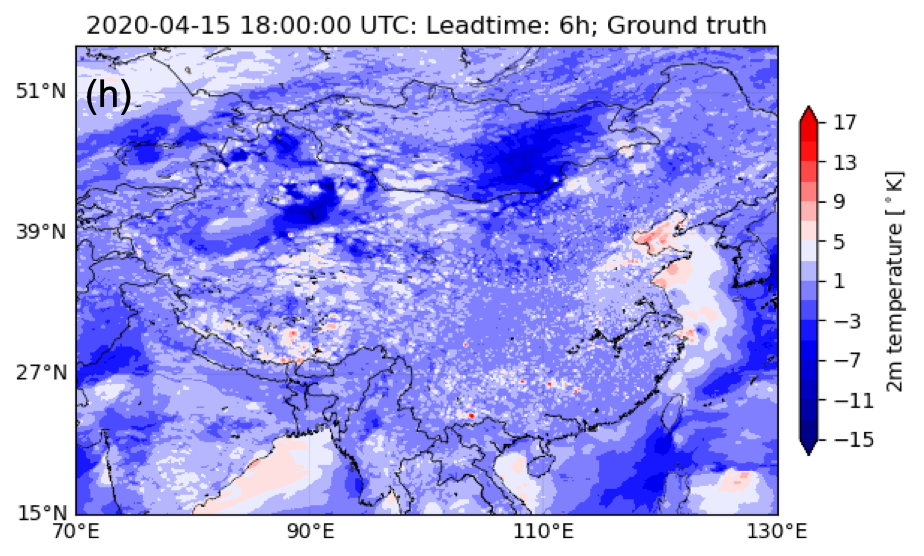}
    \end{subfigure}
    \hspace{-0.82cm}
    \begin{subfigure}[t]{0.27\textwidth}
        \includegraphics[width=\linewidth]{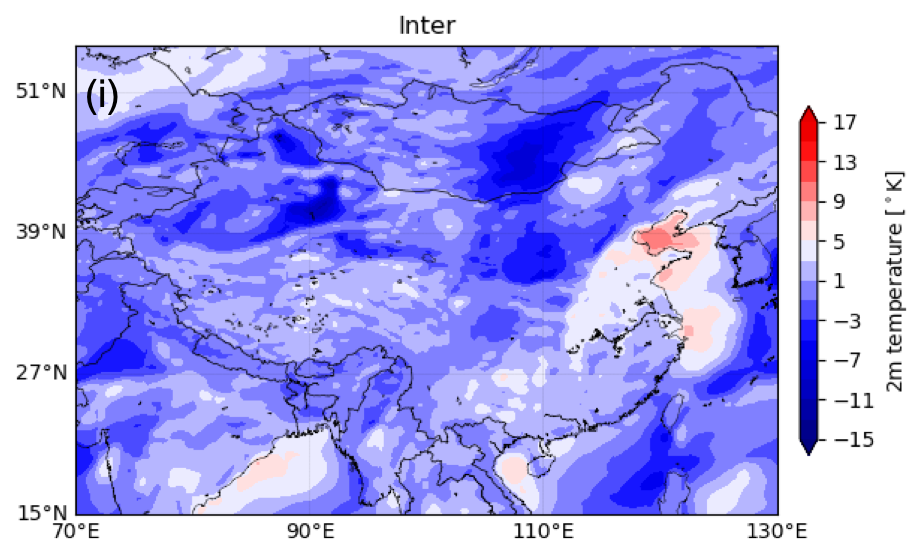}
    \end{subfigure}
     \hspace{-0.82cm}
    \begin{subfigure}[t]{0.27\textwidth}
        \includegraphics[width=\linewidth]{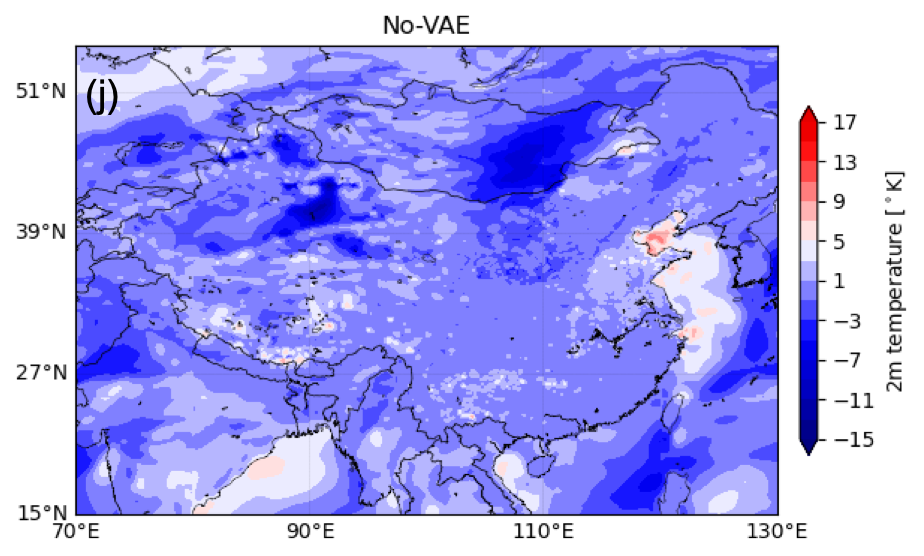}
    \end{subfigure}
    \hspace{-0.82cm}
    \begin{subfigure}[t]{0.27\textwidth}
        \includegraphics[width=\linewidth]{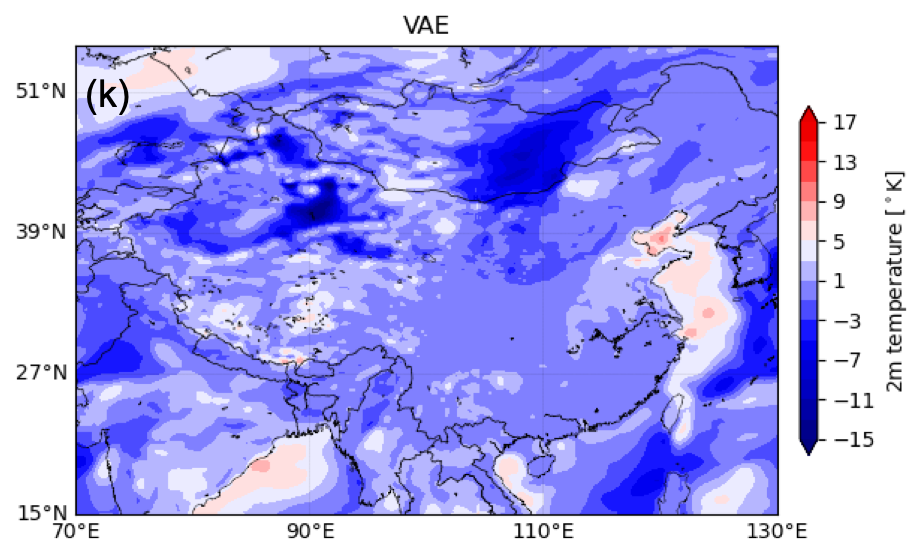}
    \end{subfigure}

    \begin{subfigure}[t]{0.27\textwidth}
        \includegraphics[width=\linewidth]{kongbai.png}
    \end{subfigure}
     \hspace{-0.82cm}
    \begin{subfigure}[t]{0.27\textwidth}
        \includegraphics[width=\linewidth]{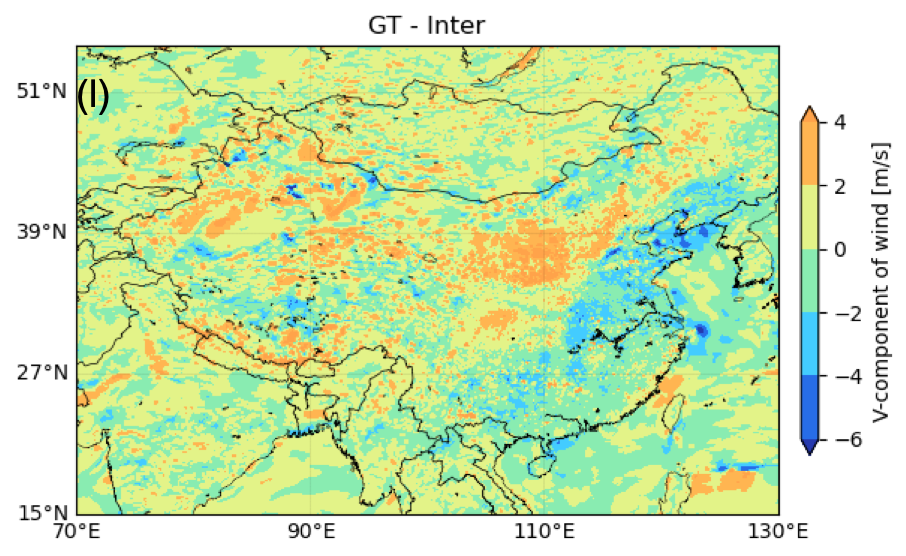}
    \end{subfigure}
         \hspace{-0.82cm}
    \begin{subfigure}[t]{0.27\textwidth}
        \includegraphics[width=\linewidth]{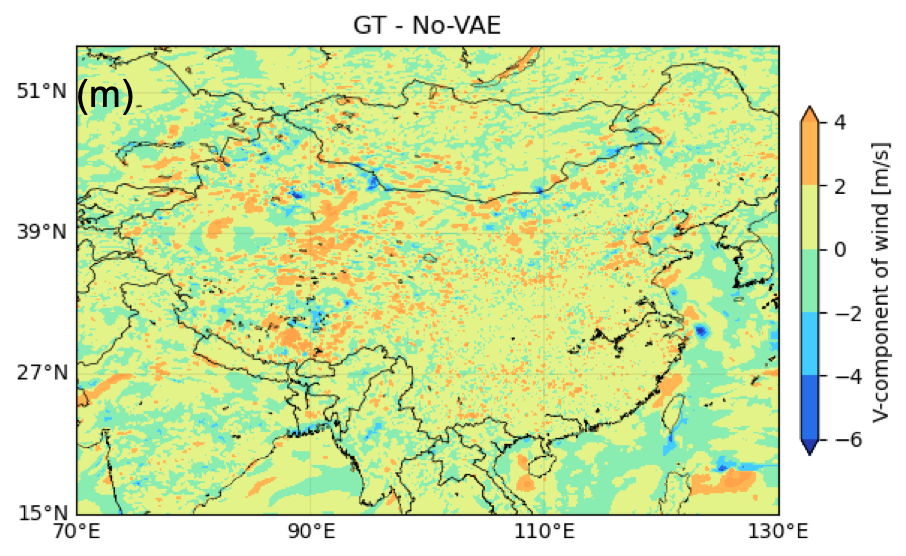}
    \end{subfigure}
    \hspace{-0.82cm}
    \begin{subfigure}[t]{0.27\textwidth} \includegraphics[width=\linewidth]{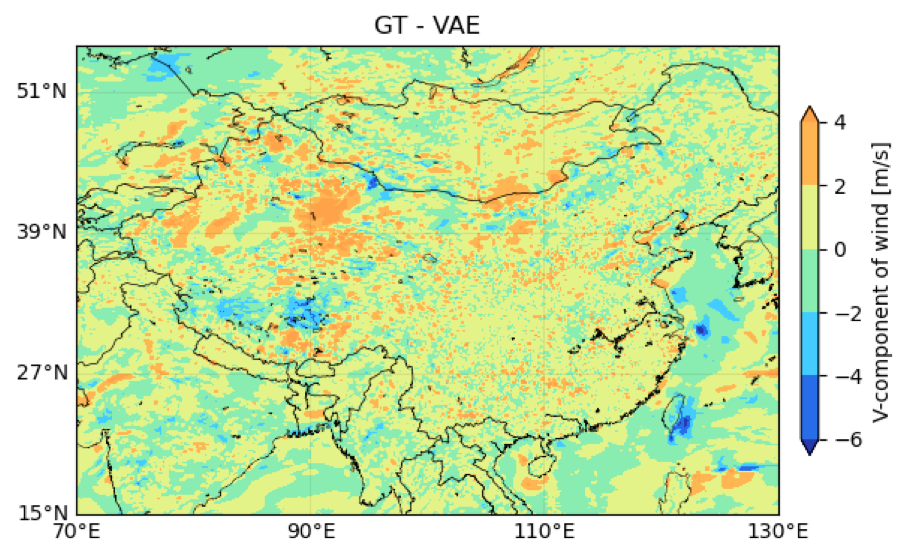}
    \end{subfigure}

    \begin{subfigure}[t]{0.27\textwidth}
        \includegraphics[width=\linewidth]{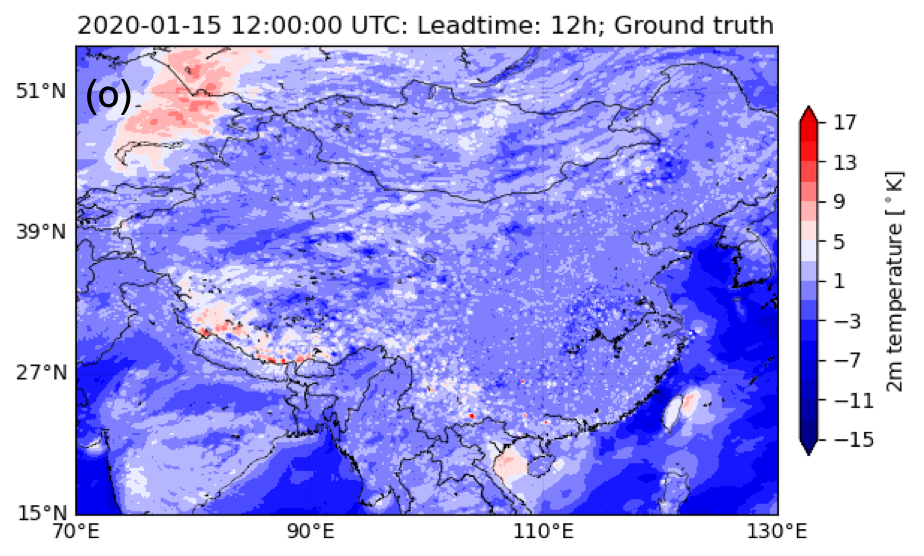}
    \end{subfigure}
    \hspace{-0.82cm}
    \begin{subfigure}[t]{0.27\textwidth}
        \includegraphics[width=\linewidth]{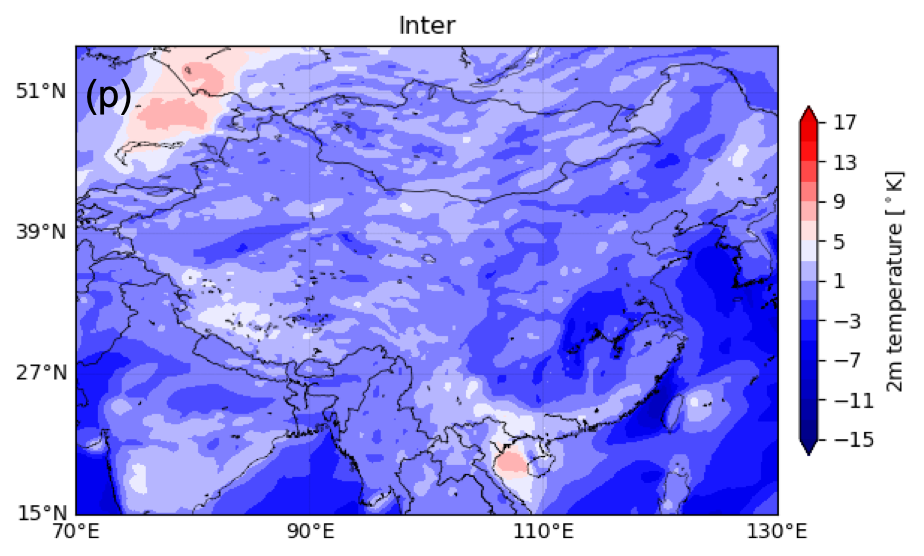}
    \end{subfigure}
     \hspace{-0.82cm}
    \begin{subfigure}[t]{0.27\textwidth}
        \includegraphics[width=\linewidth]{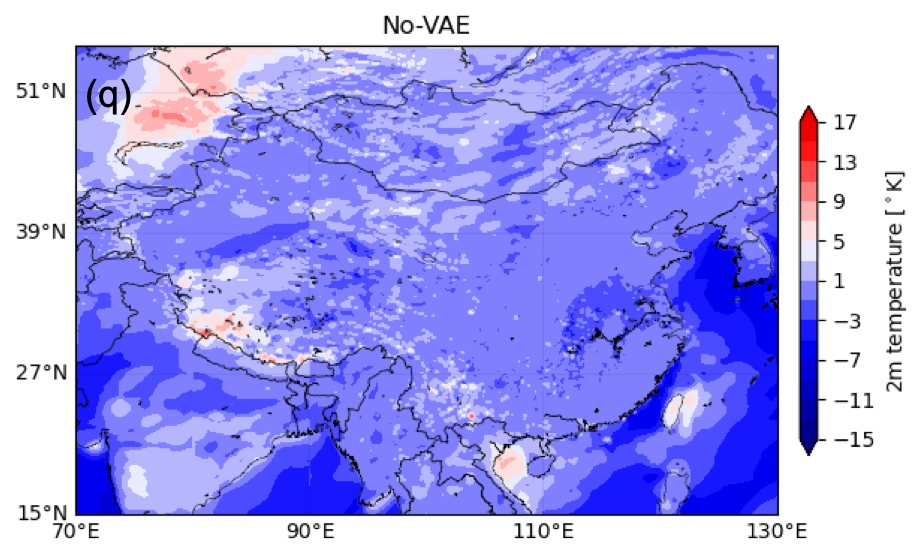}
    \end{subfigure}
    \hspace{-0.82cm}
    \begin{subfigure}[t]{0.27\textwidth}
        \includegraphics[width=\linewidth]{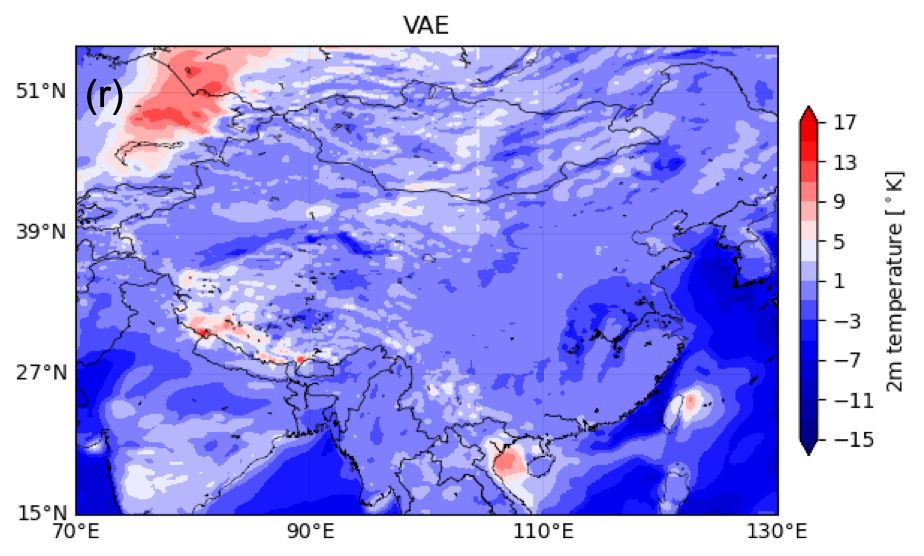}
    \end{subfigure}

    \begin{subfigure}[t]{0.27\textwidth}
        \includegraphics[width=\linewidth]{kongbai.png}
    \end{subfigure}
     \hspace{-0.82cm}
    \begin{subfigure}[t]{0.27\textwidth}
        \includegraphics[width=\linewidth]{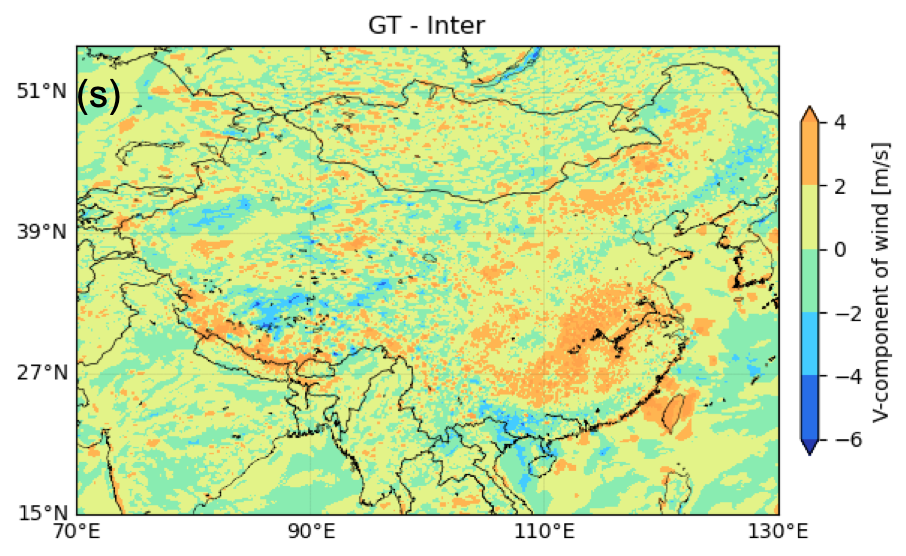}
    \end{subfigure}
         \hspace{-0.82cm}
    \begin{subfigure}[t]{0.27\textwidth}
        \includegraphics[width=\linewidth]{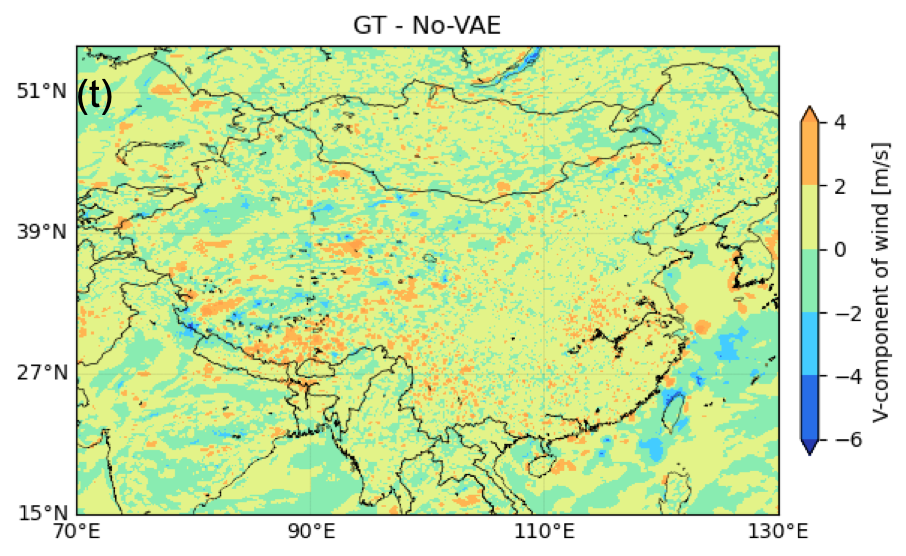}
    \end{subfigure}
    \hspace{-0.82cm}
    \begin{subfigure}[t]{0.27\textwidth} \includegraphics[width=\linewidth]{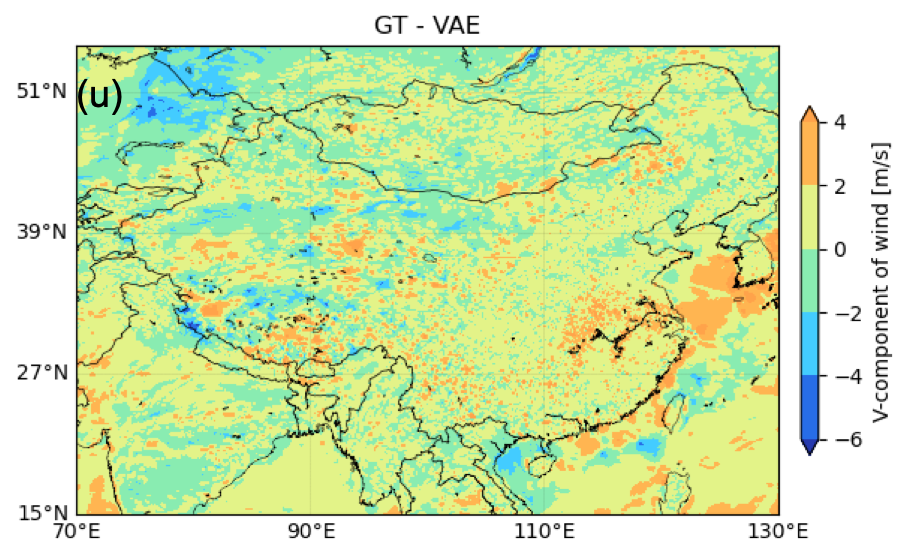}
    \end{subfigure}

    \caption{Three examples of downscaling for 10 meter V-component of wind with a lead time of 1~h (panels: b - d), 6~h (panels: i-k) and 12~h (p-f), by interpretation, No-VAE, and VAE methods respectively. (a,h,o) are the ground truth from the HRCLAS dataset. (e-g, l-n, s-u) The difference between the ground truth and the downscaled fields generated by interpretation, No-VAE, and VAE method at the time of 1\ts{st} November 2019, 01:00 UTC,  15\ts{st} April 2020, 18:00 UTC, and 15\ts{st} December 2020, 12:00 UTC. 
}
    \label{fig:sample2}
\end{figure*}

To better understand the realism of downscaling performance, we selected three examples to compare the performance of temperature and zonal wind variables using "eyeball" analysis (see Fig \ref{fig:sample1} and Fig \ref{fig:sample2}). These examples correspond to 1\ts{st} November 2019, at 01:00 UTC, 15\ts{th} April 2020, at 18:00 UTC, and 15\ts{th} January 2020, at 12:00 UTC, representing FuXi forecasts with 1-hour, 6-hour, and 12-hour lead times, respectively. As illustrated in Fig \ref{fig:sample1}, it can be seen that the interpretation method shows significant biases in temperature downscaling for southwestern China, particularly within the complex mountainous terrain of the Himalayas (see topography map in Fig~.\ref{fig:top}). This bias was especially pronounced in northeastern China during the cold winter period on 15\ts{th} January 2020, when the No-VAE interpolation methods were applied. In contrast, the application of the VAE method with the compressed data allowed the U-Net to more effectively capture and reduce errors in this region.

\section{Conclusion and Discussion}\label{sec4}

This study investigates the compression of high-dimensional, high-volume weather data using the NIC method, inspired by advancements in the computer vision domain, for downstream weather and climate applications. Specifically, we employed our proposed VAE framework to compress three years of high-resolution HRCLDAS data, reducing the data size from 8.61 TB to a compact 204 GB, an impressive 42-fold reduction in storage requirements. The results demonstrate that the VAE framework, enhanced with a fine-tuning strategy, outperforms other baseline methods, significantly reducing reconstruction errors, thereby providing more accurate reconstructions of 2-meter temperature, 10-meter U-component of wind, and 10-meter V-component of wind, in terms of RMSE. Moreover, the VAE-based model demonstrates superior performance in preserving extreme values, effectively maintaining both high and low values from the original HRCLDAS dataset.

To further validate the effectiveness and usability of the compact data generated by the VAE framework, we used it for the downscaling task. The performance metrics reveal no significant differences in MSE and SSIM between the original HRCLDAS data and compact HRCLDAS data. Although a slight degradation in performance was observed with the compact data, likely due to the reduced number of parameters utilized by the U-Net model. Further evaluation based on the power spectrum suggests that both the original HRCLDAS and compact data significantly outperform the baseline bilinear interpolation method. While using the compact data results in a minor loss of information at larger scales, it effectively retains key information at finer scales for all three target variables, similar to the original HRCLDAS data. This demonstrates the usability and effectiveness of the compact data for downstream downscaling tasks. Looking ahead, this approach has the potential to be expanded to other applications, such as large-scale weather and climate forecasting, where the need for processing extensive datasets is more crucial.

\bmhead{Code Availability Statement}
The source code used for training and running VAE compression and downscaling models in this work is available at https://doi.org/10.5281/zenodo.13862076.

\bmhead{Acknowledgements}

We extend our sincere to the researchers at ECMWF for their invaluable contributions to the collection, archival, dissemination, and maintenance of the ERA5 reanalysis dataset. We thank the China Meteorological Administration (CMA) to maintain and provide HRCLDAS data to support this research. The authors also gratefully acknowledge the computing for the Future at Fudan (CFFF) University by providing the computing resource.




\bibliography{sn-bibliography}

\end{document}